\documentclass{ecai} 

\usepackage{latexsym}
\usepackage{amssymb}
\usepackage{amsmath}
\usepackage{amsthm}
\usepackage{booktabs}
\usepackage{enumitem}
\usepackage{graphicx}
\usepackage{color}

\usepackage{algorithm}
\usepackage{algorithmic}
\usepackage{subcaption}
\usepackage{multirow}
\usepackage{booktabs}
\usepackage{amssymb}

\newcommand{\BibTeX}{B\kern-.05em{\sc i\kern-.025em b}\kern-.08em\TeX}

\begin{document}


\begin{frontmatter}


\paperid{0640} 


\title{Local Information Matters: A Rethink of Crowd Counting}


\author[A]{
\fnms{Tianhang}~\snm{Pan}\footnote{Email: panth@njust.edu.cn.}
}

\author[A]{
\fnms{Xiuyi}~\snm{Jia}\thanks{Corresponding Author. Email: jiaxy@njust.edu.cn.}
}

\address[A]{School of Computer Science and Engineering, Nanjing University of Science and Technology, Nanjing, China}


\begin{abstract}
The motivation of this paper originates from rethinking an essential characteristic of crowd counting: individuals (heads of humans) in the crowd counting task typically occupy a very small portion of the image. 
This characteristic has never been the focus of existing works: they typically use the same backbone as other visual tasks and pursue a large receptive field.
This drives us to propose a new model design principle of crowd counting: emphasizing local modeling capability of the model. 
We follow the principle and design a crowd counting model named Local Information Matters Model (LIMM). 
The main innovation lies in two strategies: a window partitioning design that applies grid windows to the model input, and a window-wise contrastive learning design to enhance the model's ability to distinguish between local density levels. 
Moreover, a global attention module is applied to the end of the model to handle the occasionally occurring large-sized individuals. 
Extensive experiments on multiple public datasets illustrate that the proposed model shows a significant improvement in local modeling capability (8.7\% in MAE on the JHU-Crowd++ high-density subset for example), without compromising its ability to count large-sized ones, which achieves state-of-the-art performance. 
Code is available at: https://github.com/tianhangpan/LIMM.
%
\end{abstract}

\end{frontmatter}


\section{Introduction}
\begin{figure}[tbp]
	\centering
    \subcaptionbox{\label{1}}{
        \includegraphics[height=75pt]{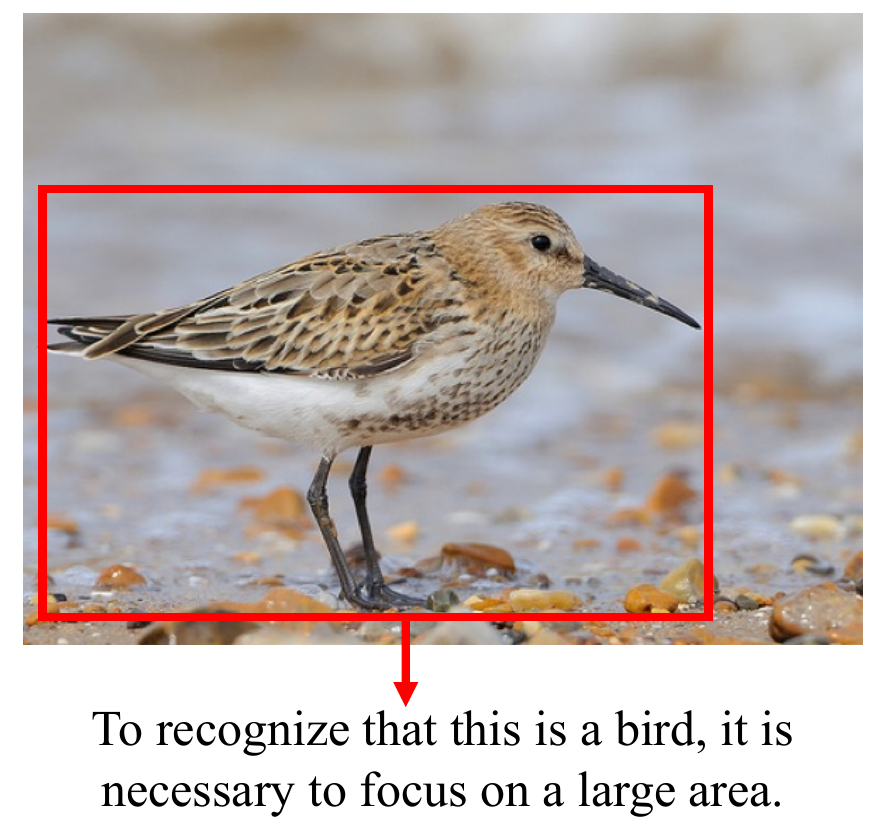}
    }
    \subcaptionbox{\label{2}}{
        \includegraphics[height=75pt]{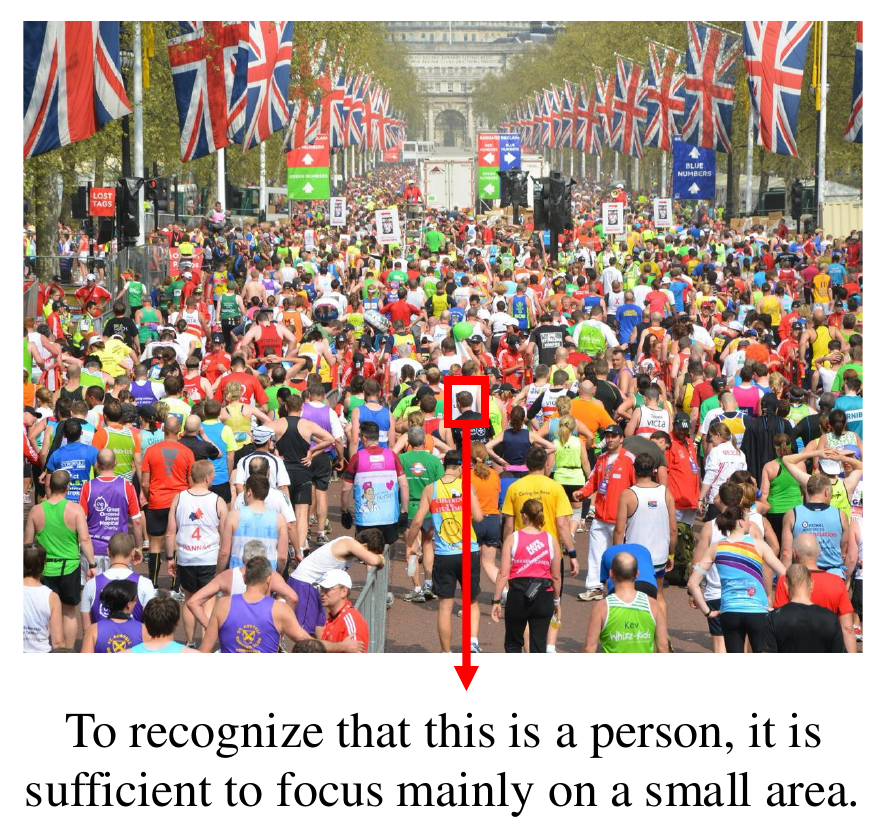}
    }
    \vspace{7pt}
	\caption[]{
 Two typical images from the ImageNet-1K image classification dataset (a) and the JHU-Crowd++ high-density crowd counting dataset (b).
 Humans exhibit distinct differences when performing these two tasks:
 attention to relatively large regions is required for image classification, whereas it is sufficient to focus on a small area when counting crowds. 
 }
    \vspace{15pt}
	\label{fig:example_imn_jhu}
\end{figure}

\begin{figure}[bp]
	\centering
    \subcaptionbox{ShanghaiTech A}{
        \includegraphics[height=45pt]{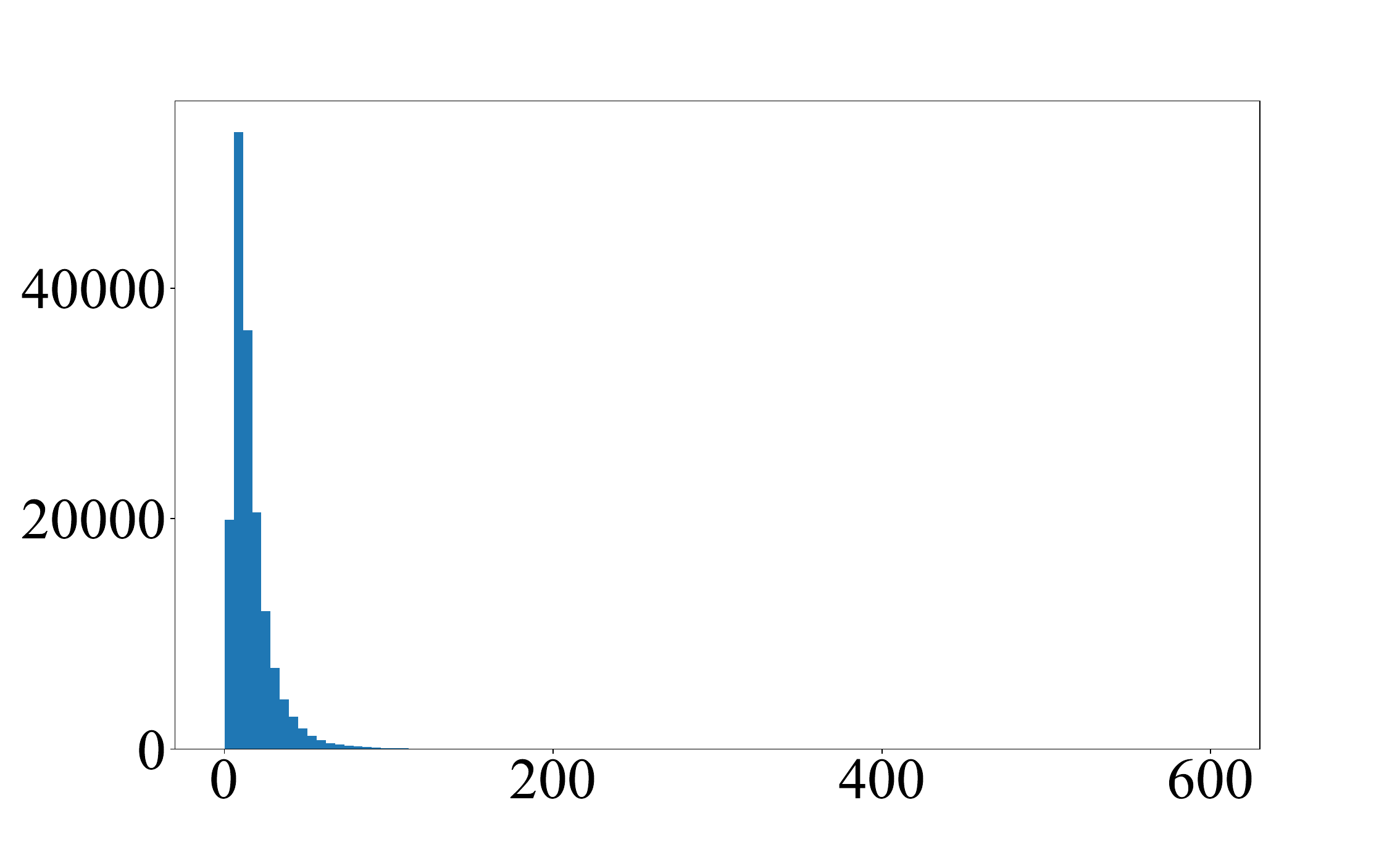}
    }
    \vspace{10pt}
    \subcaptionbox{ShanghaiTech B}{
        \includegraphics[height=45pt]{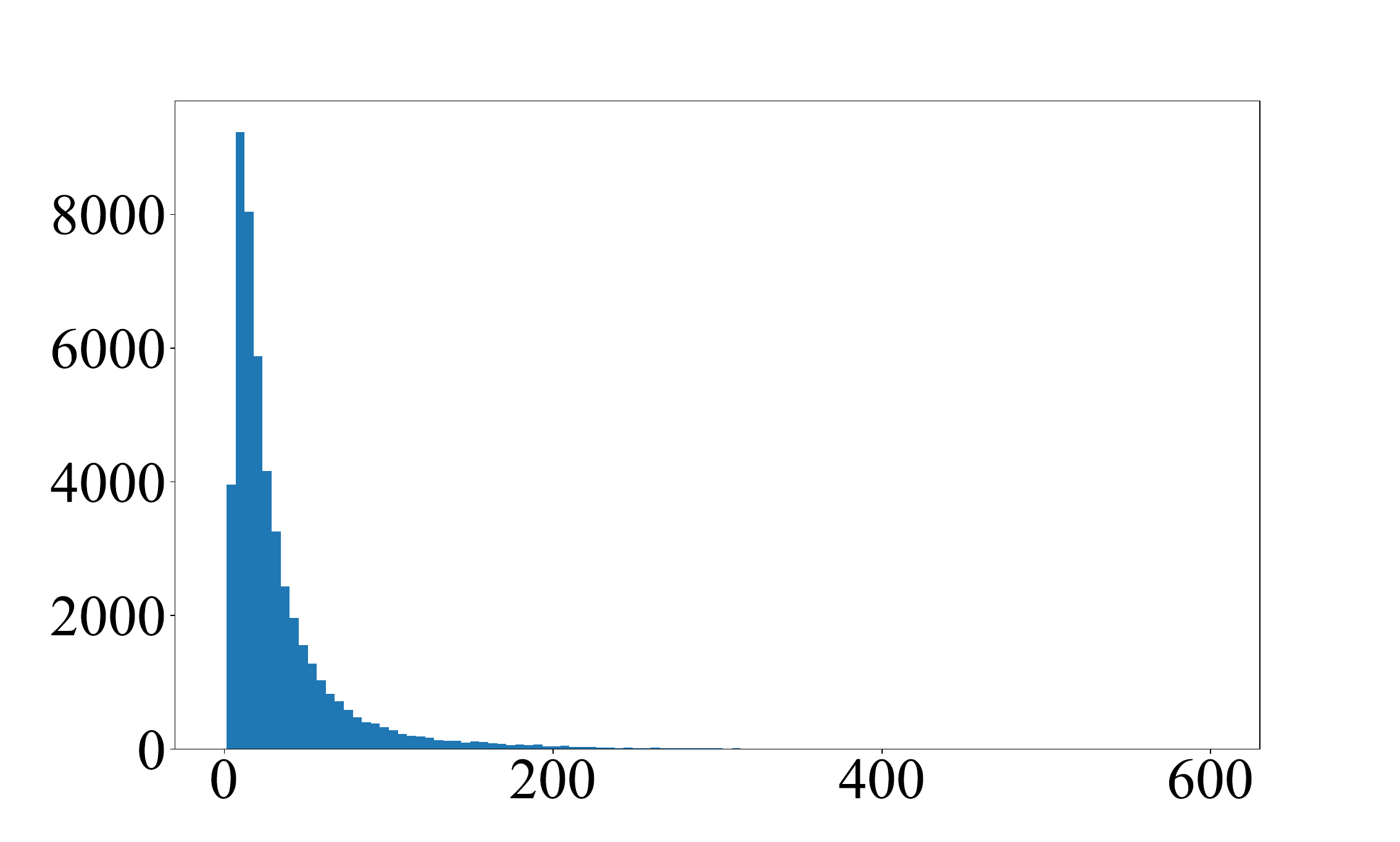}
    }
    \subcaptionbox{ImageNet-1K}{
        \includegraphics[height=45pt]{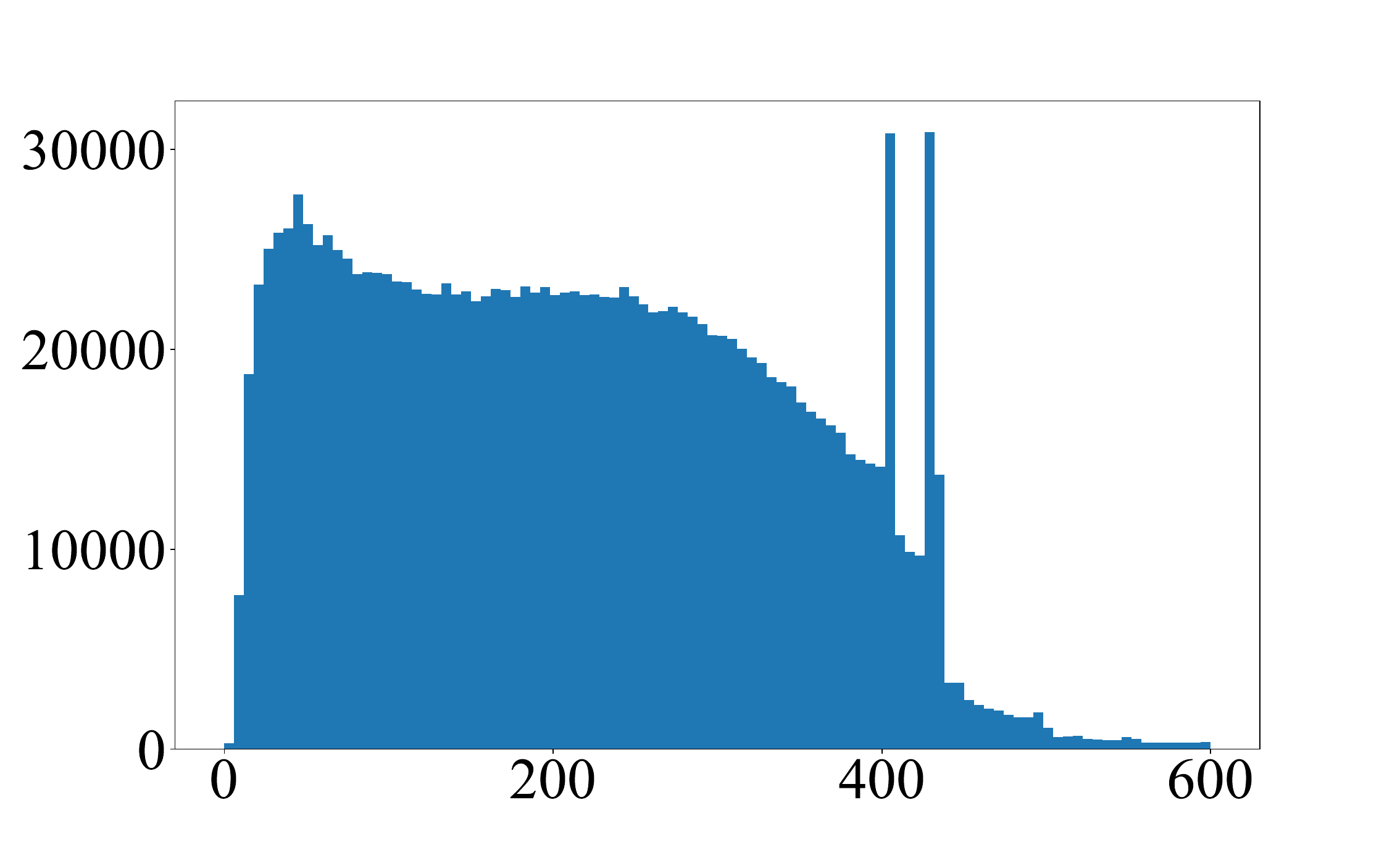}
    }
    \subcaptionbox{UCF-QNRF}{
        \includegraphics[height=45pt]{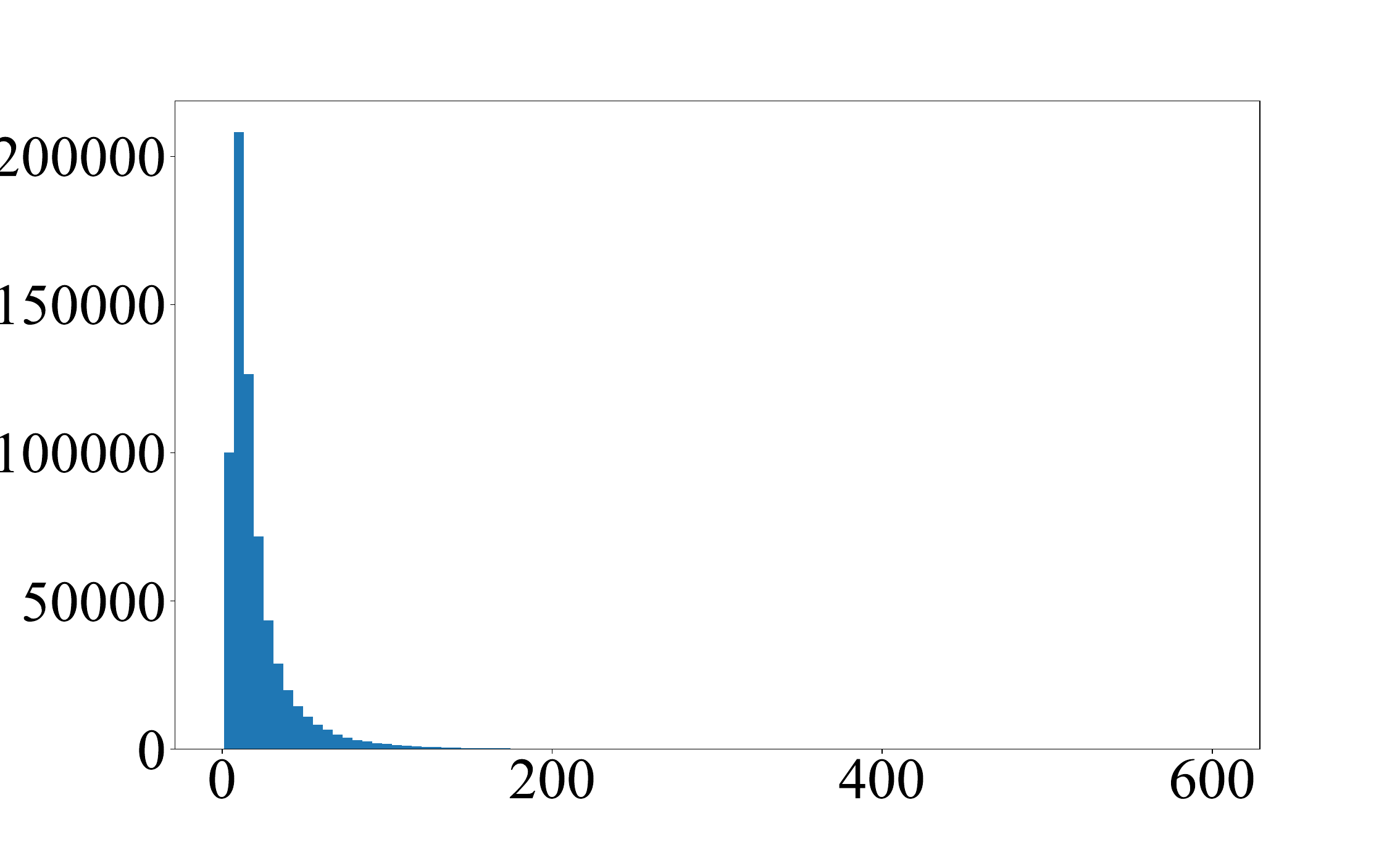}
    }
    \subcaptionbox{JHU-Crowd++}{
        \includegraphics[height=45pt]{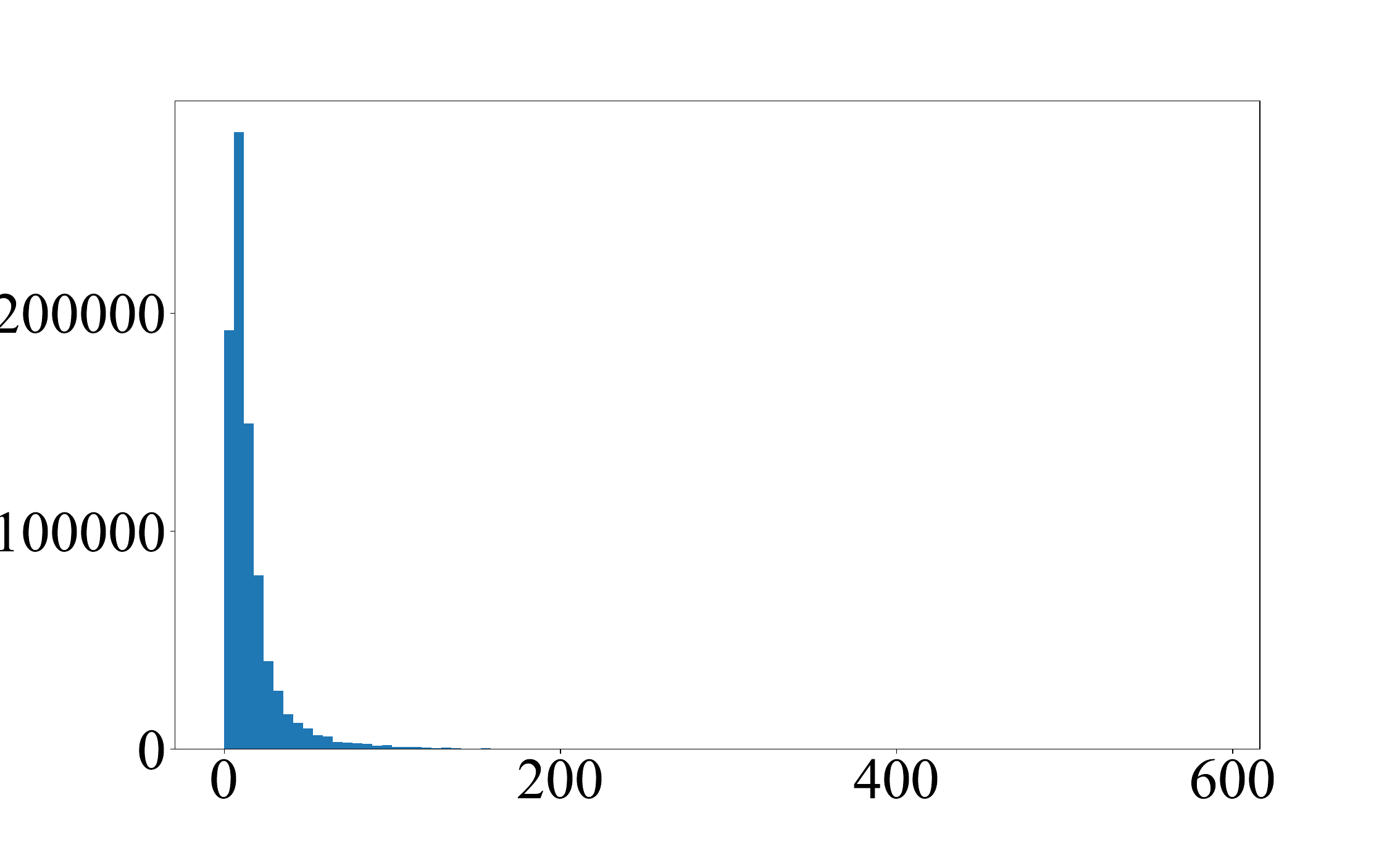}
    }
    \subcaptionbox{COCO}{
        \includegraphics[height=45pt]{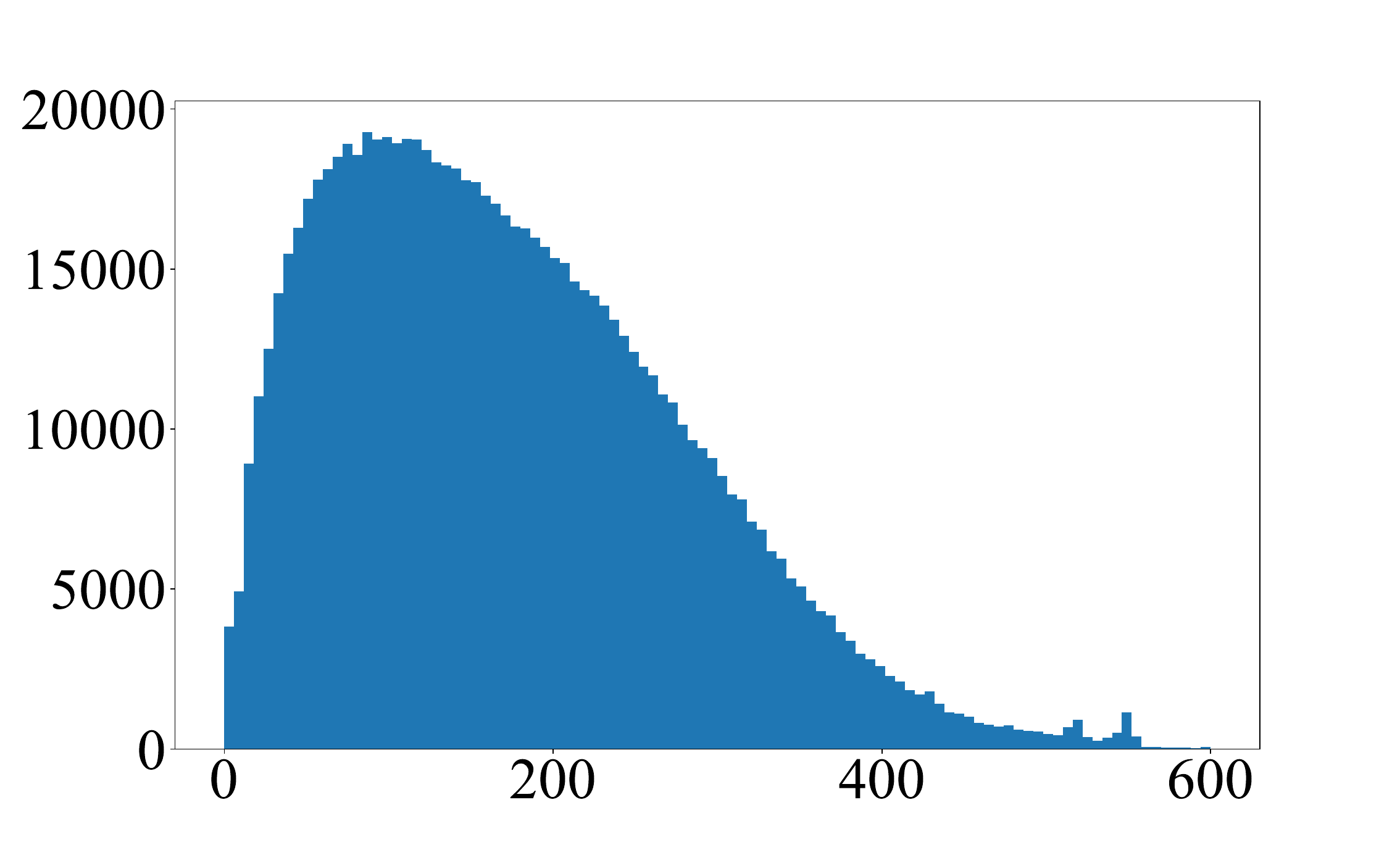}
    }
    \vspace{10pt}
	\caption[]{
 The histograms of sizes of all individuals in the training sets of datasets for different visual tasks. 
 The horizontal axis represents head size, while the vertical axis represents the count.
 }
    \vspace{15pt}
	\label{fig:size_hist}
\end{figure}

Crowd counting has attracted much attention recently because of its wide range of applications like public safety, video surveillance and traffic control~\cite{zhang2016single, ling2023motional}. 
%
%
This paper aims to explore the distinctions in characteristics between crowd counting and most other visual tasks, and to design crowd counting methods that are congruent with the unique characteristics.
Let us begin with two typical images.
Figure~\ref{fig:example_imn_jhu} (a) is a typical image from ImageNet-1K~\cite{deng2009imagenet} image classification task; 
Figure~\ref{fig:example_imn_jhu} (b) is a typical image from JHU-Crowd++~\cite{sindagi2020jhu} crowd counting task. 
Classifying figure (a) necessitate attention to a large area at a time since the bird occupies a large portion of the image space.
While due to the high density of crowd counting images, each human head typically occupies only a very small portion of the image.
To recognize a single person in figure (b), it is sufficient to focus mainly on a small area. 
The histograms of individual sizes in different tasks shown in Figure~\ref{fig:size_hist} are also strong evidences supporting this characteristic.
For example, the average head size of the JHU-Crowd++ dataset is only 16.2 pixel and heads smaller than 50 pixels account for 95.5\% of the total.
By way of contrast, the distribution of individual sizes in the ImageNet-1K image classification task~\cite{deng2009imagenet} and COCO object detection task~\cite{lin2014microsoft} differs greatly from that in crowd counting dataset. 
The method for computing the individual sizes can be found in Appendix C.
%

%
%
%

%
Contrary to the characteristics of the task, in terms of the current state of research, almost all crowd counting methods directly use backbone models designed for image classification tasks, such as VGG~\cite{simonyan2014very}, ResNet~\cite{he2016deep} and ViT~\cite{dosovitskiy2020image}. 
Even many recent crowd counting researches attempt to expand the receptive field.
For example, the dilation design in CSRNet~\cite{li2018csrnet} and the ViT backbone in CLTR~\cite{lian2021locating}. 
In our view, models with small receptive fields are already capable of handling most individuals in crowd counting tasks.
The investigation in Section~\ref{sec:investigation} can also validate this viewpoint.
We calculate and visualize the effective receptive field (ERF)~\cite{luo2016understanding} of several models trained on several visual tasks.
The results show that the ERF of models trained on the crowd counting task is not only typically much smaller than that of models trained on the image classification task and the object detection task but also significantly smaller than their theoretical receptive field (TRF)~\cite{luo2016understanding}.
Hence, simply expanding the TRF brings negligible gains in counting accuracy for small-size individuals, yet significantly increases computational burden.
We believe that designing crowd counting models should adhere to an important principle: 
\textit{emphasizing local modeling capability of the model. }
To the best of our knowledge, no previous work has taken this principle as the core of model design.
%
%
%

%

To validate the correctness of the principle, we conduct model design attempts, the final outcome is named \textbf{L}ocal \textbf{I}nformation \textbf{M}atters \textbf{M}odel (LIMM).
%
%
To align the model with the principle, we adopt a window partition design for the input image to compel the model to focus on local regions.
Experiments show that due to boundary effect, simply segmenting the input into windows does not improve model performance.
The advantage of the window partition design is realized only when a window shift strategy is introduced to mitigate the negative impact of boundary effect.
Secondly, since the windows are relatively small, during training, the windows can be categorized into different density levels according to the actual number of people within the window. 
Then multi-queue contrastive learning~\cite{pan2024boosting} can be used to enhance the model's ability to distinguish between different local density levels.
Considering that representations of similar density levels should be relatively more alike, we analysis the contrastive loss gradient and introduce queue distance weighting into multi-queue contrastive learning.
%
%
Finally, to avoid sacrificing counting accuracy for large-sized human heads, a global sub-sampled attention layer is introduced at the end of the model to supplement large-scale features.
Experiments show that after integrating the above designs, LIMM significantly improves local modeling capability without compromising accuracy for large-sized individuals, achieving state-of-the-art performance.
Furthermore, although the target of this work is to design crowd counting methods, we also acknowledge that crowd counting is not the only vision task dominated by small-sized individuals.
The experiments applying LIMM to the tiny object detection (TOD) task also reveals its potential for broader application in small-target visual tasks.
The contributions of this paper are summarized as follows:
\begin{itemize}
    \item To point out better direction for crowd counting model design, we conducted investigations into the characteristics of the data and the models. A design principle is proposed:
    \textit{emphasizing local modeling capability of the model.}
    \item To validate the correctness of the principle, we propose LIMM, which incorporates two innovative designs: window partition design and window-wise contrastive learning of density levels, into basic backbones to align with the aforementioned principle. Experiments show that LIMM achieves state-of-the-art performance on multiple crowd counting datasets.
\end{itemize}

\section{Related Work}
\subsection{Crowd Counting}
Up to now, most crowd counting methods can be divided into two categories: density map-based and localization-based methods.
Density map-based methods aim to generate a density map, the pixel values of which represent the crowd density at that location.
The sum of all pixel values in the density map is the estimated total number of the image. 
MCNN~\cite{zhang2016single} is a pioneer in employing such a method. Benefiting from the multi-column design, MCNN can handle input images of arbitrary size or resolution.
CSRNet~\cite{li2018csrnet} employs dilated CNN for the back end to deliver larger reception fields and to avoid pooling operations. 
Lin et al.~\cite{lin2022boosting} proposed a multifaceted attention network to improve transformer models in local spatial relation encoding. 
To address the difficulty to obtain high-quality density maps due to complex background interference, congested crowd and large-scale variations, Rohra et al.~\cite{rohra2025msffnet} presents a multi-scale feature fusion network which is capable of detecting enough semantic features to understand crowds in sparse and highly congested scenes. 
Lin et al.~\cite{lin2024gramformer} proposed a graph-modulated transformer to enhance the network by adjusting the attention and input node features respectively on the basis of two different types of graphs.
However, density map-based methods fail to meet the requirement of locating individual entities.
Early methods obtain the positions by post processing from bounding boxes~\cite{sam2020locate} or density maps~\cite{gao2019domain, idrees2018composition}.
More recently, Song et al.~\cite{song2021rethinking} proposed a model called P2PNet to directly predict a set of point proposals to represent heads in an image, being consistent with the human annotation results.
CLTR~\cite{liang2022end} introduced the Transformer architecture to the field of crowd localization and proposing a KMO-based Hungarian matcher to reduce the ambiguous points and generate more reasonable matching results. 
APGCC~\cite{chen2024improving} provides clear and effective guidance for proposal selection and optimization, addressing the
core issue of matching uncertainty in point-based crowd counting.
Different from above methods, our LIMM focuses on the small-scale characteristic of crowd counting, enhancing the model’s local modeling capability, and achieves state-of-the-art performance. 
This is a design perspective that previous models have not considered.

\subsection{Tiny Object Detection}
\label{sec:related_tod}
Tiny object detection (TOD) is a task sharing numerous commonalities with crowd counting.
The target of TOD is categorizing and locating tiny objects of interest in images.
In terms of scale, TOD is generally more extreme than crowd counting, as it contains almost no large-sized individuals.
The AI-TOD dataset~\cite{wang2021tiny} is a typical example, with its images captured via aerial photography.
The largest instance in AI-TOD is smaller than 64 pixels, and more than 86\% of the instances are smaller than 16 pixels. 
TinyPerson~\cite{yu2020scale} is a dataset for human detection, which opens up a promising direction for tiny object detection in a long distance and with massive backgrounds. 
In the TinyPerson dataset, 73\% of the instances are smaller than 20 pixels.
Approaches to address the TOD task are diverse, e.g., focusing attention on objects~\cite{li2020cross}, image super-resolution~\cite{rabbi2020small}, and modeling the relationship between the environment and the objects~\cite{torralba2003contextual}.
Although TOD fundamentally differs from crowd counting, we experimentally apply LIMM to TOD in Section~\ref{sec:tiny_obj} to further validate its performance in handling extremely small-sized targets.

\section{ERF Investigation into Crowd Counting Models}
\label{sec:investigation}
In this section, ERF and TRF~\cite{luo2016understanding} are utilized to investigate the attention scope of crowd counting models. 
ERF~\cite{luo2016understanding} refers to the size of the input image region that has a non-negligible impact on the output unit, and it can reflect whether the model focuses more on local or global regions when solving the task.
ERF can be represented by a gradient map. 
The gradient of each pixel $g_{i,j}$ is the partial derivative of the output unit $y_{p,q}$ with respect to the input pixel $x_{i,j}$, which can be calculated by back-propagating. 
%
%
Any pixel with an impact gradient value exceeding 1\%–95.45\% of the center point is considered as in the ERF. 
%
%
In our investigation, the input is the image itself.
In the feature maps output from the final layer of the backbone model, we select the maximum channel value at the center point as the output.
\begin{figure*}[tbp]
	\centering
	\includegraphics[width=380pt]{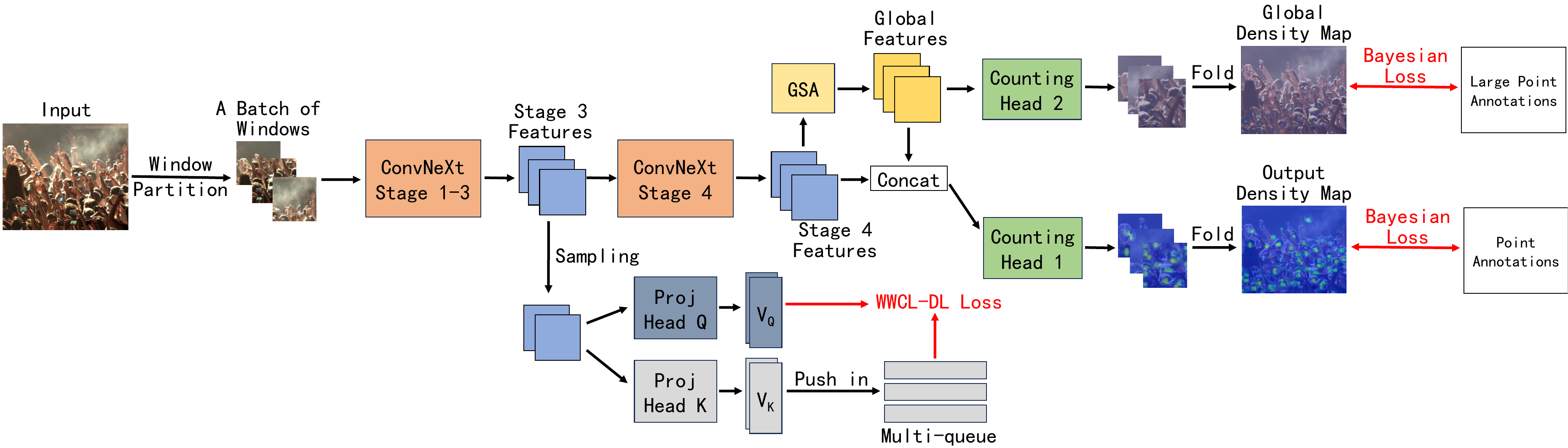}
	\caption[]{
 The training structure of LIMM. 
 WWCL-DL Loss refers to the loss of window-wise contrastive learning of density levels. 
 }
    \vspace{5pt}
	\label{fig:structure}
\end{figure*}
\begin{figure}[tbp]
	\centering
    \includegraphics[width=200pt]{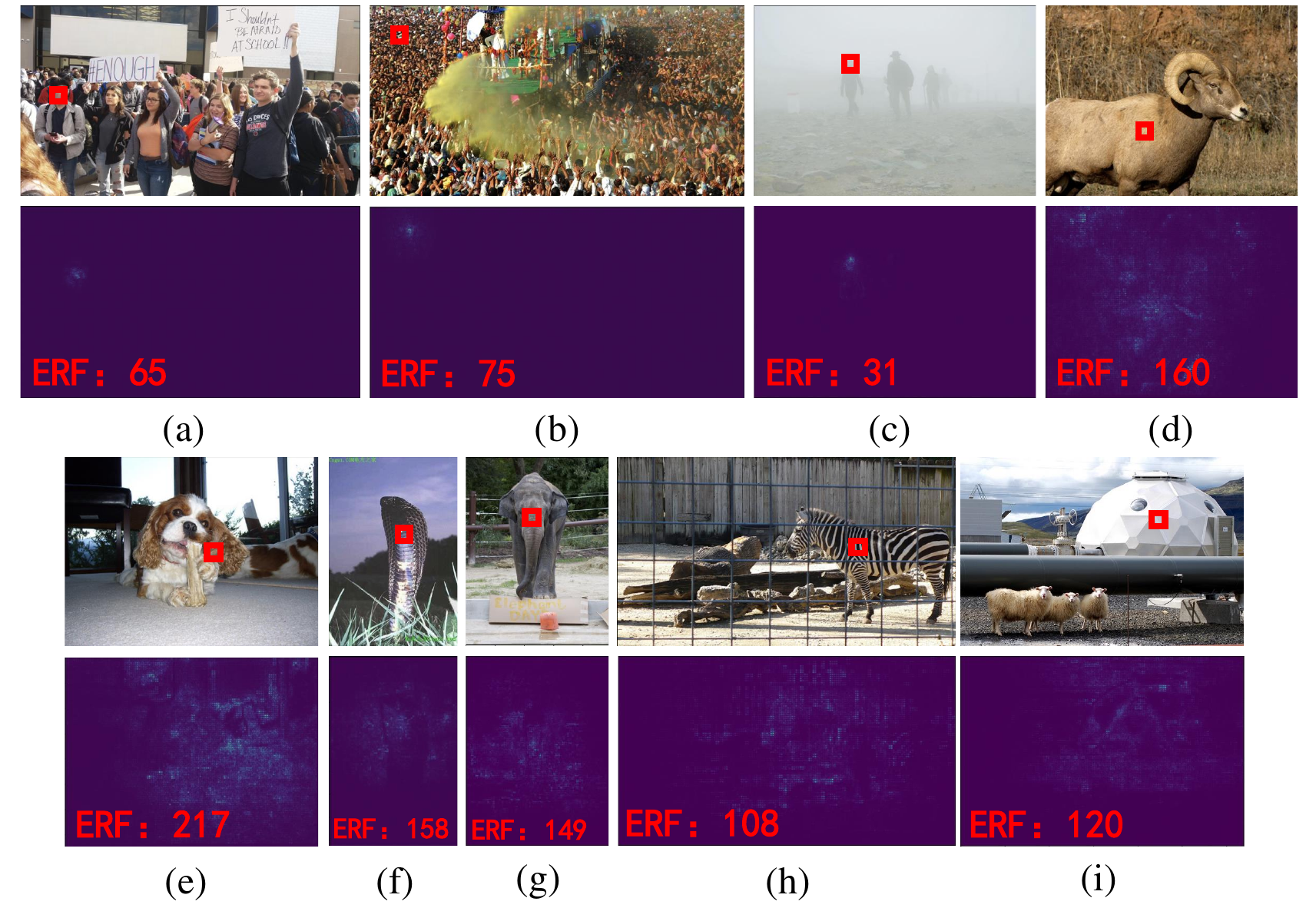}
	\caption[]{
 Visualization of the ERF gradient map. 
 Figures (a) - (c) are selected from the JHU-Crowd++ dataset, Figures (d) - (f) are selected from the ImageNet-1K dataset and Figures (g) - (i) are selected from the COCO dataset.
 The models are ConvNeXt-T, trained separately on these two datasets.
 The red boxes indicate the positions of the center points.
 }
	\label{fig:erf_visual}
\end{figure}
We train several models separately on the JHU-Crowd++ crowd counting dataset, the ImageNet-1K image classification dataset and the COCO object detection dataset.
Note that when trained on different tasks, the models corresponding to the same backbone differ only in the task-specific head, with the rest of the structure being identical.
The structure of the heads is introduced in Appendix B.
The training settings is explained in Section~\ref{sec:set_up}.
We select some images from the three datasets and visualize the ERF gradient maps generated by the ConvNeXt models, as shown in Figure~\ref{fig:erf_visual}.
Obviously, the ERF region of the crowd counting models is generally concentrated around the head region of an individual, and is significantly smaller than the ERF of image classification models and object detection models.
In addition, quantitative analysis is performed to verify the generality of this phenomenon.
Specifically, on each image from the training set of the datasets, five center points are randomly sampled to calculate the ERF size. 
Finally, the average of the ERF sizes corresponding to all sampled center points is taken as the quantitative result.
As an additional reference, we introduce the conception of theoretical receptive field (TRF)~\cite{luo2016understanding}.
TRF refers to the size of the input image region that has neural connections to the output value. 
%
%
%
%
We follow the method of the work~\cite{koutini2019receptive} to calculate TRF. 
The results of Table~\ref{tbl:erf_trf_comp} show that the average ERF size of the crowd counting models is significantly smaller than that of other models. 
Even in object detection tasks, due to the frequent occurrence of large-sized individuals, models do not solely concentrate on local regions.
Moreover, the significant difference between the ERF and TRF of crowd counting models indicates that such a large TRF design is generally redundant.
In summary, in contrast to most other visual tasks, focusing on local information is sufficient to handle most cases in crowd counting. 
\begin{table}[tbp]
	\centering
	\tiny
    \caption{The average ERF and TRF of several backbone models trained on the JHU-Crowd++ dataset, the ImageNet-1K dataset and the COCO dataset. }
	\begin{tabular}[]{ccccc}
		\toprule
		Models & JHU ERF & ImageNet ERF & COCO ERF & TRF \\
		\midrule
		CSRNet~\cite{li2018csrnet} & 72 & 144 & 149 & 284 \\
        ConvNeXt-T~\cite{liu2022convnet} & 87 & 156 & 139 & 1688 \\
        Swin-T~\cite{liu2021swin} & 91 & 238 & 194 & Full image \\
        MAN~\cite{lin2022boosting} & 76 & 233 & 203 & Full image  \\
		\bottomrule
	\end{tabular}
	\label{tbl:erf_trf_comp}
\end{table}
%

%
%

%
\begin{figure}[tbp]
	\centering
    \subcaptionbox{\label{1}}{
        \includegraphics[height=80pt]{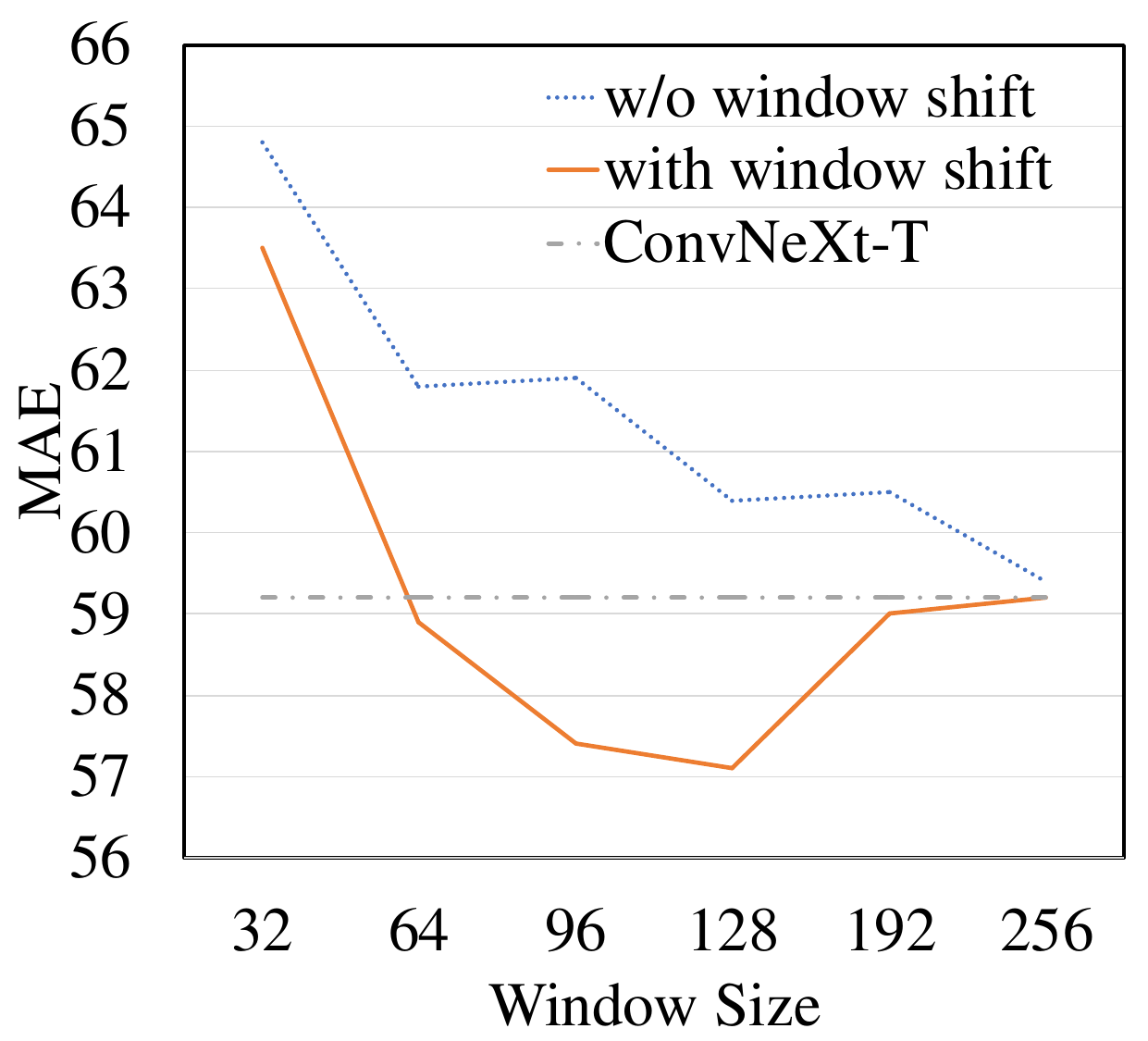}
    }
    \subcaptionbox{\label{2}}{
        \includegraphics[height=80pt]{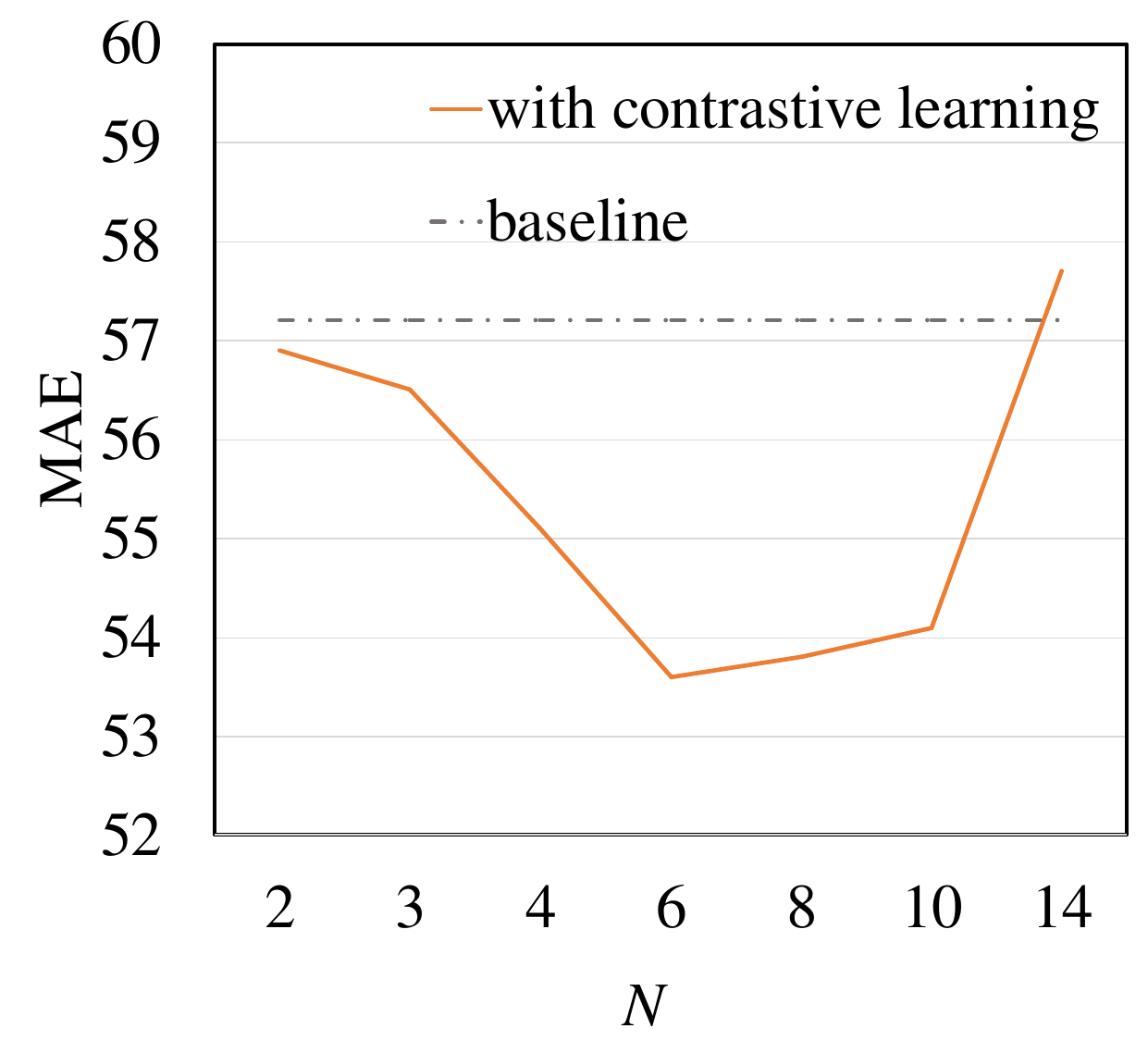}
    }
    \vspace{10pt}
	\caption[]{
 (a) Model performance under different window size settings.
 (b) Model performance under different number of density levels $N$.
 The baseline is ConvNeXt-T with the window partition design and the expirments are conducted on the JHU-Crowd++ dataset. 
 }
    \vspace{15pt}
	\label{fig:ws_ndl}
\end{figure}

\section{Local Information Matters Model (LIMM)}
Based on the investigation and analysis, we believe that designing crowd counting models should adhere to the following principle:
\textit{emphasizing local modeling capability of the model.}
To verify whether the principle can effectively guide the design of excellent crowd counting models, we attempt a model design, the final outcome is named \textbf{L}ocal \textbf{I}nformation \textbf{M}atters \textbf{M}odel (LIMM).
Regarding the backbone model, we believe that CNNs with strong local priors align better with the principle. 
Therefore, ConvNeXt~\cite{liu2022convnet} is regarded as the preferred choice for the backbone model.
The training structure of LIMM is shown in Figure~\ref{fig:structure}.

\subsection{Window Partition Design}
\begin{figure}[tbp]
	\centering
	\includegraphics[height=80pt]{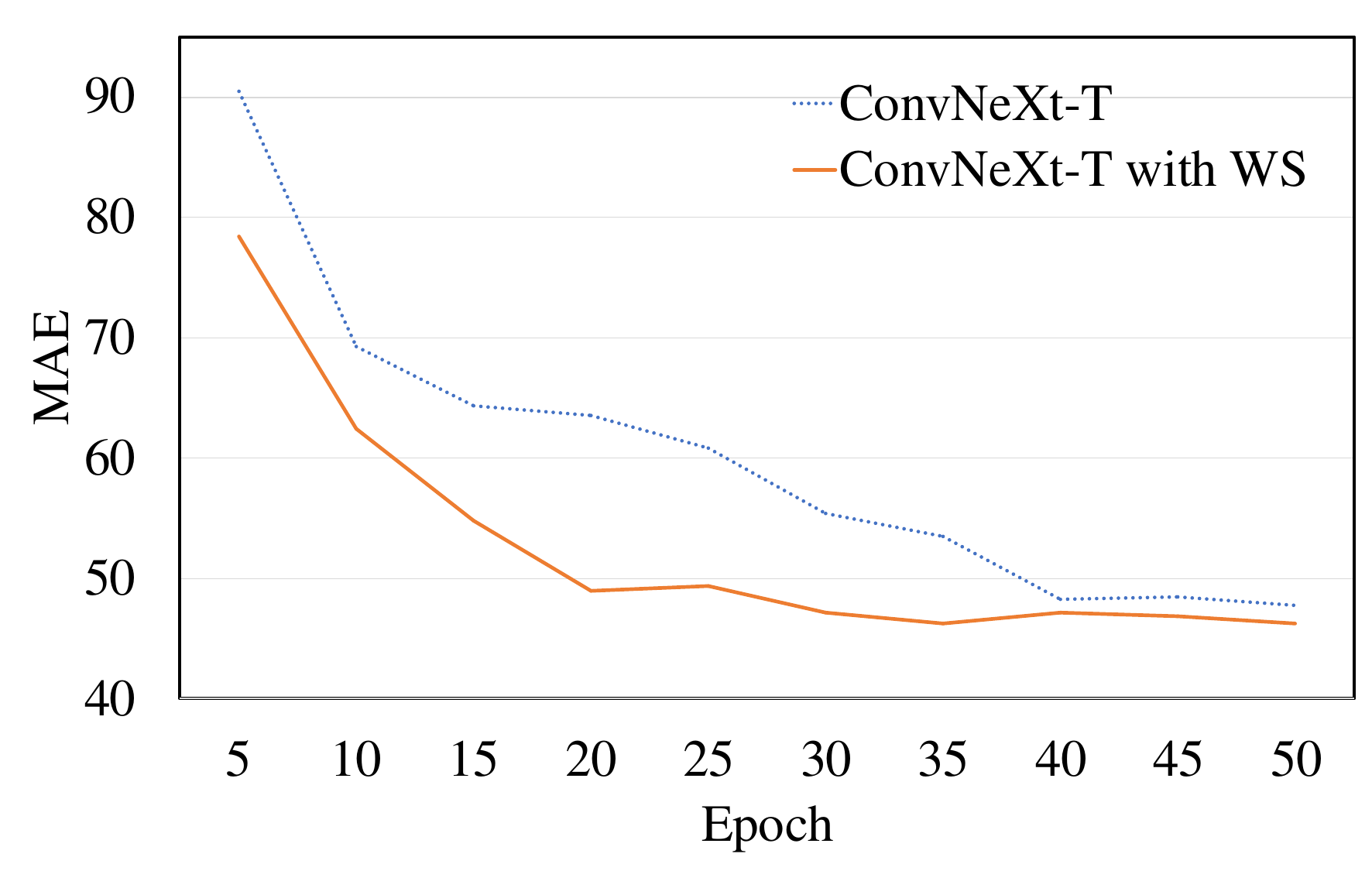}
	\caption[]{
 The validation accuracy during the training of the model with window partition design on the JHU-Crowd++ dataset. 
 Compared to baseline, the model with window partition design can converge in fewer epochs and achieve better performance after convergence.
 }
    \vspace{15pt}
	\label{fig:train_speed}
\end{figure}

Contrary to the principle, even the smallest Tiny version of the ConvNeXt model family has a huge TRF of 1688.
This indicates that it is designed to model information from relatively large regions.
To align the model with the principle, we attempt a simple method of partitioning the input image into small grid windows, limiting the model's TRF in the window. 
These windows are fed into the model in parallel, and the output density maps are eventually folded into a complete density map.
Experiment on the JHU-Crowd++ dataset is conducted to investigate the impact of varying window sizes on model performance.
The results are shown in Figure~\ref{fig:ws_ndl} (a).
Unfortunately, simply applying window partition design to the input image has led to a performance degradation.
This is because the boundaries of the windows can divide individuals, increasing the difficulty for the model to recognize them. 
We refer to this as the boundary effect.
To address this issue, we employ a shift-window mechanism similar to that of Swin~\cite{liu2021swin} in the second block of the first stage of ConvNeXt-T.
Since window shift is only applied once, the model can obtain information nearby the window edges without significantly increasing the TRF.
As shown in Figure~\ref{fig:ws_ndl} (a), after introducing window shift, the method with window partition design demonstrated significant improvement compared to the baseline ConvNeXt.
According to the experiments from Section~\ref{sec:investigation}, models without window partition design can also inherently learn to focus on local regions. 
However, our window partition design can still significantly enhance the model's performance.
It can be attributed to the following rationale: window partitioning eliminates the model's ability to capture long-distance information, reducing the solution space and allowing the model to converge more quickly to the correct solution.
Figure~\ref{fig:train_speed} shows the change in validation accuracy during the training of the model with window partition design.
Obviously, compared to the baseline, it can converge in fewer epochs and achieve better performance after convergence.
Among the different window sizes, as shown in Figure~\ref{fig:ws_ndl} (a), the moderate size of 128 is chosen as the default value due to its superior performance.
If the window is too small (e.g., 32), it will frequently fail to cover a single individual, leading to significant information loss within the window and causing substantial difficulties in counting.
If the window is too large (e.g., 128), although it does not cause significant performance degradation, the model will be quite similar to the original ConvNeXt, which does not align with the principle of emphasizing local modeling capability.

\subsection{Window-wise Contrastive Learning of Density Levels}
A model with excellent local modeling capability should be able to encode the content of the windows into good representations:
1) be able to accurately distinguish between foreground and background regions within the windows; 
2) be able to accurately represent the density of the crowd within the window.
Fortunately, because the window is very localized, we can treat each window as an individual sample. 
During training, we sample $S$ windows in each image and categorize the windows into different density levels based on the true number of people within the window (background is also considered a density level), and utilize multi-queue contrastive learning~\cite{pan2024boosting} to enhance the model's ability to differentiate between density levels.
A detailed description of multi-queue contrastive learning is provided in Appendix D.
In this paper, the feature maps output by the third stage of ConvNeXt are used for contrastive learning.
Samples with the same density level are treated as positive examples, while those with different density levels are considered negative examples.
Additionally, determining the thresholds for defining different density levels is also a challenge.
We aim to balance the number of instances from different classes as much as possible during contrastive learning.
Therefore, before training, extensive random sampling of 128-sized windows are performed on the training set of each dataset. 
We then set quantiles evenly based on the number of density levels $N$ to obtain the density classification thresholds.
For example, a total of 3000 samples were taken. 
After sampling, the samples were sorted based on the actual number of people within each window. 
If the number of density levels $N$ is 3, the number of people in the 1000-th sample and the 2000-th sample are taken as the density classification thresholds, respectively.
As shown in Figure~\ref{fig:ws_ndl} (b), above contrastive learning strategy can indeed significantly improve the performance. 
The number of density levels $N$ is chosen to be 6, which achieved the best performance.
If a smaller $N$ (e.g., 3) is chosen, the model can accurately distinguish between foreground and background but lacks the ability to finely represent density.
Conversely, a larger $N$ (e.g., 14) causes windows with very similar densities to be treated as negative samples and pushed apart, which is detrimental to the model's learning.
To further explore the characteristics of the window size and the number density levels, we also investigate their optimal values on three more datasets, as outlined in Appendix F.
The above contrastive learning strategy has already significantly improved model performance. 
However, the situation in this paper differs from most other contrastive learning scenarios: the degree of association between different classes varies.
For example, the distance between the representations of windows with density level 1 and density level 2 should be smaller compared to the distance between density level 1 and density level 3.
Our approach is to reduce the repulsion between samples with similar density levels during the contrastive learning process, while increasing the repulsion between samples with more dissimilar density levels.
To achieve this, we attempt to analyze the gradients of negative examples in multi-queue contrastive learning, where the magnitude of the gradient reflects the strength of mutual repulsion between negative examples.
The contrastive loss of a batch of vectors $V_{Q}$ is:
\begin{equation}
    L_{C} = \frac{1}{|N_{Q}|} L_{i}, 
    \label{eq:lc}
\end{equation}
where $N_{Q}$ is the size of the batch and $L_{i}$ is the loss corresponding to vector $V_{Qi}$: 
\begin{equation}
    L_{i} = \frac{-1}{|P_{i}|} \sum_{p \in P_{i}} \log{\frac{\exp(s_{i,p} / \tau)}{\sum_{j=1}^{J} \exp(s_{i,j} / \tau)}}, 
\end{equation}
where $P_{i}$ is the set of samples in the queue corresponding to the class of sample $V_{Qi}$, $s_{i,p}$ is the cosine similarity between $V_{Qi}$ and $V_{Kp}$.
$V_{Kp}$ is the vector stored in the multi-queue structure, $\tau$ is the temperature coefficient. 
Assuming $V_{Qi}$ and $V_{Kk}$ do not belong to the same class, the partial derivative of $L_{i}$ with respect to $s_{i,k}$ is:
\begin{equation}
    \frac{\partial L_{i}}{\partial s_{i,k}} = \frac{1}{\tau} \frac{\exp(s_{i,k} / \tau)}{\sum_{j=1}^{J} \exp(s_{i,j} / \tau)}. 
\end{equation}
The magnitude of it reflects the strength of mutual repulsion between negative examples.
We envisage a loss function $L^{'}_{i}$ to replace $L_{i}$ in Equation~\ref{eq:lc}, the partial derivative of it differs from $\partial L_{i} / \partial s_{i,k}$ by an additional weight $w_{i,k}$, the value of which is:
\begin{equation}
    w_{i,k} = 
    \left\{
    \begin{array}{cc}
        0, & d_{i,k} = 0  \\
        f(x), & d_{i,k} = x, x \geq 1
    \end{array}, 
    \right.
\end{equation}
where $d_{i,j}$ is the class distance between $V_{Qi}$ and $V_{Kk}$.
$f(x)$ needs to be positively correlated with $x$ to achieve a larger partial derivative for negative classes that are farther apart.
The partial derivative of $\partial L^{'}_{i} / \partial s_{i,k}$ is:
\begin{equation}
    \frac{\partial L^{'}_{i}}{\partial s_{i,k}} = \frac{w_{i,k}}{\tau} \frac{\exp(s_{i,k} / \tau)}{\sum_{j=1}^{J} \exp(s_{i,j} / \tau)}. 
\end{equation}
$L^{'}_{i}$ be obtained by integrating it:
\begin{equation}
    L^{'}_{i} = \frac{-1}{|P_{i}|} \sum_{p \in P_{i}} \log{\frac{w_{i,j} \exp(s_{i,p} / \tau)}{\sum_{j=1}^{J} w_{i,j} \exp(s_{i,j} / \tau)}}.
\end{equation}
Various settings of $f(x)$ are tried, and the performance comparisons are shown in Table~\ref{tbl:fx_comp}.
The three settings of $f(x)$ that are positively correlated with $x$ all outperform the baseline $f(x) = 1$. 
$f(x) = 2^x$ outperforms the linear and power functions due to its greater contrast.
Finally, the loss of window-wise contrastive learning of density levels (WWCL-DL) is:
\begin{equation}
    L_{\rm WWCL-DL} = \frac{1}{|N_{Q}|} L^{'}_{i}. 
\end{equation}
\begin{table}[tbp]
	\centering
	\tiny
    \caption{Comparisons of different $f(x)$ in contrastive learning. }
	\begin{tabular}[]{ccc}
		\toprule
		$f(x)$ & MAE & RMSE \\
		\midrule
        $f(x) = 1$ & 53.6 & 229.5 \\
		$f(x) = x$ & 53.7 & 227.5 \\
        $f(x) = x^2$ & 53.4 & 226.3 \\
        $f(x) = 2^x$ & \textbf{53.0} & \textbf{219.9} \\
		\bottomrule
	\end{tabular}
	\label{tbl:fx_comp}
\end{table}

\subsection{Feature Supplement for Large-sized Individuals}
After introducing window partition design and contrastive learning, the model's ability to model local information has been significantly improved.
However, the performance improvement comes with some trade-offs.
There are still a small number of large-sized human heads in the data, exceeding the size of a single window, which makes detection challenging.
Although the inaccuracy in counting these individuals does not significantly affect the performance on the entire test set, it represents a decline in the model's robustness.
To address the issue, we introduce a global sub-sampled attention (GSA) layer~\cite{chu2021twins} to supplement the final features of the backbone model.
GSA uses a single key to summarize the important information for each of window and the key is used to communicate with other windows.
It takes the features from the fourth stage output of ConvNeXt as input, and its output is concatenated with the original features.
As a result, information from different windows can be integrated, allowing, for example, the GSA layer to recognize that parts such as the eyes and mouth, located in different windows, belong to the same face.
To train the GSA layer, another crowd counting head is added after its output to generate a global density map. 
Since we only want the GSA layer to capture features related to large-sized heads, we supervise it using only large point annotations with head size greater than 50 pixel.
The model structure of this sub-section is shown in the upper-right part of Figure~\ref{fig:structure}.
In summary, the total loss function of LIMM is:
\begin{equation}
    L = L_{\rm FB} + \lambda_1 * L_{\rm GB} + \lambda_2 * L_{\rm WWCL-DL}, 
\end{equation}
where $L_{\rm FB}$ is the Bayesian Loss~\cite{ma2019bayesian} that supervises the output density map, $L_{\rm GB}$ is the Bayesian Loss that supervises the global density map and $L_{\rm WWCL-DL}$ is the loss of our window-wise contrastive learning of density levels. 
During inference, we retain only the window partition design, the ConvNeXt backbone, the GSA layer and the counting head 1.
The sum of pixel values in the output density map is treated as the predicted number. 
\begin{table*}[tbp]
	\centering
	\tiny
    \caption[]{Quantitative results comparing with the state-of-the-art methods on four datasets. The best performance is shown in \textbf{bold} and the second best is shown in \underline{underlined}. ``-" indicates that public performance data is unavailable. ``$\downarrow$" indicates that the smaller the value, the better the performance.}
	\begin{tabular}[]{ccccccccc}
		\toprule
		Dataset & \multicolumn{2}{c}{ShanghaiTech A} & \multicolumn{2}{c}{ShanghaiTech B} & \multicolumn{2}{c}{UCF-QNRF} & 
		\multicolumn{2}{c}{JHU-Crowd++} \\
        \midrule
		Method & MAE $\downarrow$ & RMSE $\downarrow$ & MAE $\downarrow$ & RMSE $\downarrow$ & MAE $\downarrow$ &
		RMSE $\downarrow$ & MAE $\downarrow$ & RMSE $\downarrow$ \\
		\midrule
		MCNN~\cite{zhang2016single} (CVPR 16) & 110.2 & 173.2 & 26.4 & 41.3 & 277.0 & 426.0 & 188.9 & 483.4 \\
        CP-CNN~\cite{sindagi2017generating} (ICCV 17) & 73.6 & 106.4 & 20.1 & 30.1 & - & - & - & - \\
        CSRNet~\cite{li2018csrnet} (CVPR 18) & 68.2 & 115.0 & 10.6 & 16.0 & - & - & 85.9 & 309.2 \\
        SANet~\cite{cao2018scale} (ECCV 18) & 67.0 & 104.5 & 8.4 & 13.6 & - & - & 91.1 & 320.4 \\
        CA-Net~\cite{liu2019context} (CVPR 19) & 61.3 & 100.0 & 7.8 & 12.2 & 107.0 & 183.0 & 100.1 & 314.0 \\
        BL~\cite{ma2019bayesian} (ICCV 19) & 62.8 & 101.8 & 7.7 & 12.7 & 88.7 & 154.8 & 75.0 & 299.9 \\
        CG-DRCN-CC~\cite{sindagi2020jhu} (TPAMI 20) & 60.2 & 94.0 & 7.5 & 12.1 & 95.5 & 164.3 & 71.0 & 278.6 \\
        DPN-IPSM~\cite{ma2020learning} (ACMMM 20) & 58.1 & 91.7 & 6.5 & 10.1 & 84.7 & 147.2 & - & - \\
        UOT~\cite{ma2021learning} (AAAI 21) & 58.1 & 95.9 & 6.5 & 10.2 & 83.3 & 142.3 & 60.5 & 252.7 \\
        P2PNet~\cite{song2021rethinking} (ICCV 21) & 52.7 & 85.1 & \underline{6.3} & \underline{9.9} & 85.3 & 154.5 & 60.3 & 238.3 \\
        MAN~\cite{lin2022boosting} (CVPR 22) & 56.8 & 90.3 & - & - & 77.3 & 131.5 & 53.4 & \underline{209.9} \\
        CLTR~\cite{liang2022end} (CVPR 22) & 56.9 & 95.2 & 6.5 & 10.6 & 85.8 & 141.3 & 59.4 & 235.2 \\
        STEERER~\cite{han2023steerer} (ICCV 23) & 55.6 & 87.3 & 6.8 & 10.7 & \underline{76.7} & 135.1 & 55.4 & 221.4 \\
        CLIP-EBC~\cite{ma2024clip} (arXiv 24) & 52.5 & 85.9 & 6.6 & 10.5 & 80.3 & 139.3 & - & - \\
        FGENet~\cite{ma2024fgenet} (MMM 24) & 51.7 & 85.0 & \underline{6.3} & 10.5 & 85.2 & 158.8 & - & - \\
        CLM + ConvNeXt-T~\cite{chen2024learning} (TIP 24) & 56.2 & 89.9 & 6.5 & 11.0 & 84.1 & 144.4 & 55.8 & 237.3 \\
        Gramformer~\cite{lin2024gramformer} (AAAI 24) & 54.7 & 87.1 & - & - & \underline{76.7} & \underline{129.5} & \underline{53.1} & 228.1 \\
        APGCC~\cite{chen2024improving} (ECCV 24) & \textbf{48.8} & \textbf{76.7} & \textbf{5.6} & \textbf{8.7} & 80.1 & 136.6 & 54.3 & 225.9 \\
        MSFFNet~\cite{rohra2025msffnet} (PAA 25) & 58.6 & 93.2 & 6.2 & 10.1 & 85.3 & 146.7 & - & - \\
		\midrule
        Swin-T~\cite{liu2021swin} (ICCV 21) & 60.6 & 91.6 & 7.4 & 12.4 & 89.9 & 154.0 & 60.4 & 268.9 \\
        LIMM + Swin-T & 52.2 & 84.6 & 6.8 & 10.2 & 78.5 & 133.6 & 55.7 & 214.1 \\
        ResNet-50~\cite{he2016deep} (CVPR 16) & 78.3 & 114.1 & 10.8 & 19.2 & 106.0 & 244.5 & 74.8 & 274.2 \\
        LIMM + ResNet-50 & 59.2 & 89.0 & 6.9 & 12.4 & 88.9 & 135.1 & 58.7 & 241.2 \\
		ConvNeXt-T~\cite{liu2022convnet} (CVPR 22) & 60.4 & 92.0 & 7.3 & 13.0 & 87.9 & 150.3 & 59.2 & 249.3 \\
        LIMM + ConvNeXt-T & \underline{50.8} & \underline{84.2} & 6.5 & 10.2 & \textbf{76.4} & \textbf{125.3} & \textbf{53.0} & \textbf{207.9} \\
		\bottomrule
	\end{tabular}
	\label{tbl:quantitative_results}
\end{table*}

\section{Experiment and Discussion}
\subsection{Experiment Setup}
\label{sec:set_up}
The experiments are conducted on four public crowd counting datasets:
ShanghaiTech Part A~\cite{zhang2016single}, ShanghaiTech Part B~\cite{zhang2016single}, UCF-QNRF~\cite{idrees2018composition} and JHU-Crowd++~\cite{sindagi2020jhu}. 
They cover a variety of crowd counting scenarios with different levels of density.
Detailed descriptions of these datasets are provided in Appendix H.
Note that we do not conduct experiment on the NWPU-Crowd~\cite{wang2020nwpu} dataset since the online evaluation website has been inaccessible since the start of this work, and it is the only way to assess the method’s performance.
Following the convention of existing works~\cite{li2018csrnet, lin2022boosting}, we adopt Mean Absolute Error (MAE) and Root Mean Squared Error (RMSE) as the metrics to evaluate the methods.
The geometry-adaptive kernel method~\cite{zhang2016single} is adopted in our method to generate the ground-truth density map.
For data augmentation in all the datasets, random scaling with a factor ranging from 0.75 to 1.25, random image cropping to a size of 512 × 512 and random horizontal flipping with a probability of 0.5 are applied. 
AdamW algorithm~\cite{loshchilov2017decoupled} with a learning rate 10e-4 is adopted, the weight decay is set as 10e-3 and the batch size is set as 16.
%
%
The number of density levels $N$ is set to 6, the length of each sub-queue $L$ is set to 1024, the function $f(x)$ in the class distance weight $d_{i,j}$ is set as $f(x) = 2^x$, the number of heads in the GSA layer is set as 8 and the number of sampled windows $S$ in each image is set to 5. 
The coefficients $\lambda_1$ and $\lambda_2$ in the loss function are set to 1 and 10, respectively.
\begin{table*}[tbp]
	\centering
	\tiny
    \caption[]{Ablation study. The table shows the MAE on the JHU-Crowd++ and the ShanghaiTech Part B dataset. WP, WWCL-DL and GSA stands for the window partition design, the window-wise contrastive learning of density levels and the global sub-sampled attention layer, respectively. }
	\begin{tabular}[]{ccc|ccccccc}
		\toprule
		WP & WWCL-DL & GSA & ShanghaiTech A & ShanghaiTech B & UCF-QNRF & JHU Low & JHU Med & JHU High & JHU Total \\
		\midrule
		& & & 60.4 & 7.3 & 87.9 & 9.2 & 33.6 & 250.5 & 59.3 \\
		\checkmark & & & 56.3 & 7.4 & 81.4 & 10.5 & 31.5 & 247.9 & 57.9 \\
		& \checkmark & & 53.0 & 7.0 & 83.1 & 9.2 & 30.5 & 240.2 & 55.9 \\
		\checkmark & \checkmark & & 52.4 & 7.0 & 78.2 & 10.1 & \textbf{28.6} & 228.8 & 53.3 \\
        \checkmark & & \checkmark & 55.9 & 7.1 & 80.1 & 9.1 & 31.4 & 247.4 & 57.5 \\
        \checkmark & \checkmark & \checkmark & \textbf{50.8} & \textbf{6.5} & \textbf{76.4} & \textbf{8.9} & \textbf{28.6} & \textbf{228.6} & \textbf{53.0} \\
		\bottomrule
	\end{tabular}
    \vspace{5pt}
	\label{tbl:ablation_study}
\end{table*}

\subsection{Comparison with State-of-the-art Methods}
We evaluate our method on the above four datasets and list 19 recent state-of-the-art methods for comparison. 
LIMM achieves great counting accuracy on all four benchmark datasets.
Note that ConvNeXt-T is our baseline.
Compared to the original version designed for image classification, we only replaced the classification head with a crowd counting head.
By comparing with baseline, LIMM shows significant improvement across all four datasets.
For example, the MAE/RMSE is decreased by 10.5\%/10.7\% on the JHU-Crowd++ dataset compared to ConvNeXt-T. 
Compared to the latest state-of-the-art methods, such as Gramformer and APGCC, LIMM outperforms them on both the UCF-QNRF and the JHU-Crowd++ dataset.
%
%
CLM is also a contrastive learning-based method.
For fair comparison, CLM in Table~\ref{tbl:quantitative_results} uses the same backbone as LIMM.
Its performance lags significantly behind LIMM because it uses only a single pixel point as a sample, thus lacking the local information within the window during training. 
Moreover, it categorizes samples into only two classes: foreground and background, without training the model to distinguish between different density levels.
In relatively low-density datasets, such as ShanghaiTech Part B, the proportion of larger-sized human-heads tends to be higher. 
LIMM's advantages are relatively less apparent in such datasets, slightly trailing behind the performance of APGCC. 
However, it still shows a 11.0\%/21.5\% performance improvement on MAE/RMSE compared to the baseline model.
To validate the adaptability of LIMM to different backbones, we have also conducted experiments by incorporating Swin-T~\cite{liu2021swin} and ResNet-50~\cite{he2016deep}, as the backbones for LIMM, respectively.
Since both are four-stage models, they can be fully integrated into the LIMM design, much like ConvNeXt-T.
The baseline models, Swin-T and ResNet-50, apart from the head structure, the rest of their architectures remain identical to the original models used for image classification tasks.
As shown in the lower part of Table~\ref{tbl:quantitative_results}, LIMM with Swin-T and ResNet-50 as backbones also demonstrates significantly superior performance across all datasets compared to their respective baselines.

\subsection{Ablation Study}
We conduct ablation study on the all the four datasets.
%
%
In addition, we separately analyzed the performance of high, medium, and low-density images in the JHU-Crowd++ test set.
Since the size of individuals in an image is typically negatively correlated with density, we use this approach to investigate how different designs in LIMM affect the counting accuracy of individuals of varying sizes.
Table~\ref{tbl:ablation_study} presents the result of ablation study.
We start with the baseline ConvNeXt-T. 
Taking the JHU-Crowd++ dataset as an example, when only the window partition design is applied to ConvNeXt, the model's prior aligns better with the task. 
However, due to the model's lack of strong density level differentiation capability within the windows, there is only a slight improvement in MAE by 2.4\% on the JHU-Crowd++ dataset.
Moreover, since the windows cannot fully cover large-sized individuals and there is no effective method to model the information between windows, the accuracy actually decreases on low-density images in the JHU-Crowd++ dataset and the ShanghaiTech Part B dataset.
If only the contrastive learning strategy is applied, the model's ability to distinguish between foreground and background as well as its density estimation capability will improve, with MAE improving by 5.7\% on the JHU-Crowd++ dataset. 
However, due to the lack of focus on local details, the model's performance has not yet reached its optimal state.
When window partition and contrastive learning strategies are applied simultaneously, the model aligns well with the principles proposed in this paper. 
It effectively focuses on local information while possessing strong foreground-background distinction and density estimation capability, resulting in a significant improvement of 10.6\% in MAE on the JHU-Crowd++ dataset.
However, at this point, the model's accuracy on low-density images in the JHU-Crowd++ dataset is still lower than that of the baseline.
After adding the GSA layer, the model's performance has significantly improved on both the JHU low-density images and the ShanghaiTech Part B dataset.
This indicates that the model now exhibits strong robustness to large-sized individuals.
This also confirms that the designs proposed in this paper are complementary and achieve their maximum effectiveness when used together.
\begin{figure}[tbp]
	\centering
	\includegraphics[width=200pt]{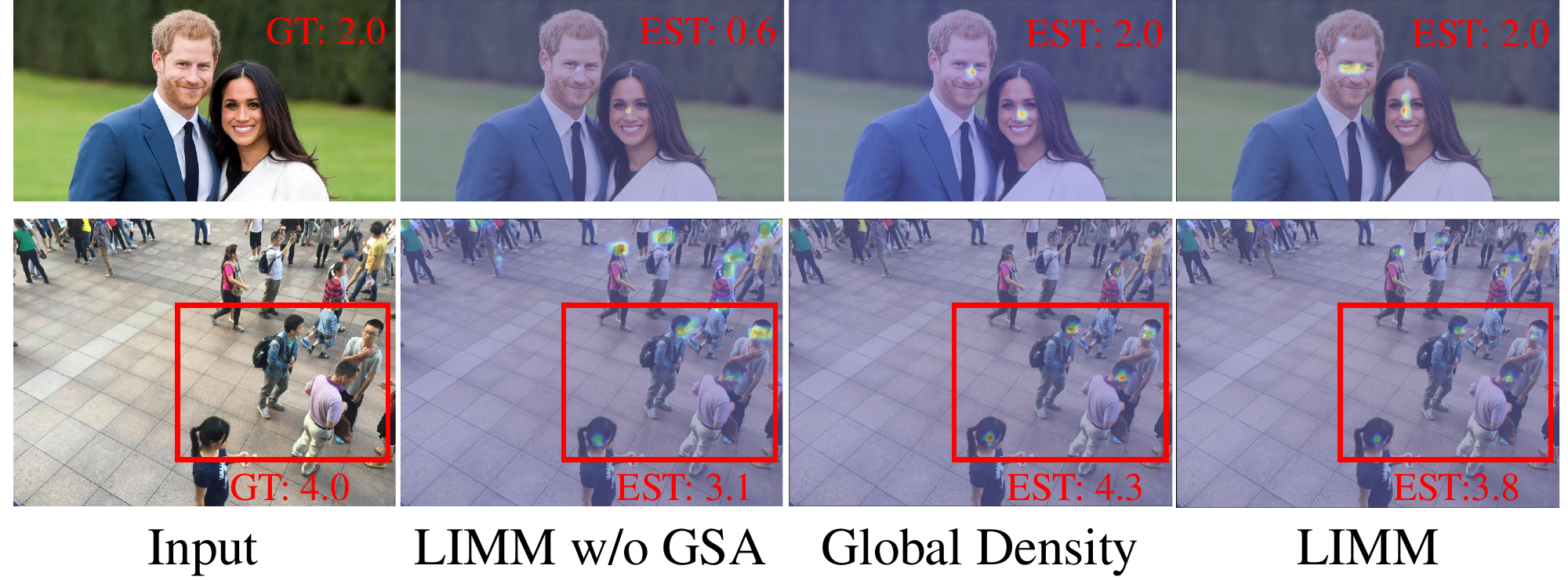}
	\caption[]{
 The visualization of the impact of the GSA layer. 
 The two input images at the top and bottom are from the JHU-Crowd++ and ShanghaiTech Part B dataset, respectively.
 }
    \vspace{15pt}
	\label{fig:gsa_visual}
\end{figure}
\begin{figure*}[tbp]
	\centering
    \includegraphics[width=380pt]{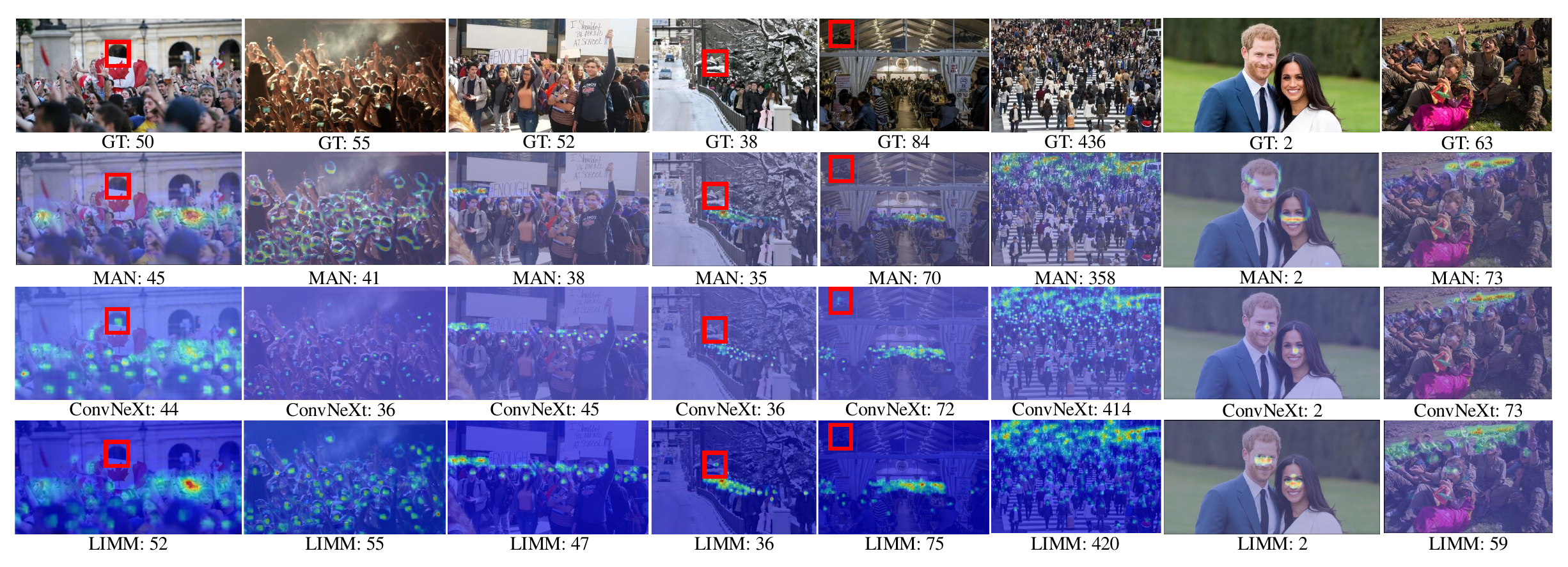}
	\caption[]{Visualization of density maps generated by LIMM, the baseline ConvNeXt and the SOTA method MAN~\cite{lin2022boosting}. Images are selected from the JHU-Crowd++ dataset~\cite{sindagi2020jhu}. GT stands for ground truth number of people.}
    \vspace{10pt}
	\label{fig:density_visual}
\end{figure*}

\subsection{Effectiveness of The GSA Layer}
In this sub-section, we visualize the impact of the GSA layer on the density maps predicted by LIMM.
As shown in Figure~\ref{fig:gsa_visual}, without the GSA layer, LIMM is unable to recognize large-sized individuals, and its accuracy is lower than that of the baseline model.
The global density map accurately identifies large-sized individuals while being less sensitive to small-sized ones. 
This indicates that the features generated by the GSA layer effectively complement the backbone features of LIMM.
The complete LIMM demonstrates strong recognition ability for individuals of all sizes.
%



%
\begin{table}[tbp]
	\centering
	\tiny
    \caption{Comparison of computational complexity and the number of parameters. The computational complexity is measured by FLOPs when inferencing images with the size of 384 $\times$ 384. 
	}
	\begin{tabular}[]{ccc}
		\toprule
		Method & FLOPs & \#param \\
		\midrule
		CLTR~\cite{liang2022end} (ECCV 22) & 37.0G & 43M \\ 
		MAN~\cite{lin2022boosting} (CVPR 22) & 58.2G & 31M \\
        CLM~\cite{chen2024learning} (TIP 24) & 28.5G & 30M \\
        Gramformer~\cite{lin2024gramformer} (AAAI 24) & 60.9G & 29M \\
		\midrule
		ConvNeXt-T~\cite{liu2022convnet} (CVPR 22) & 27.0G & 29M \\
		LIMM + ConvNeXt-T & 27.5G & 30M \\
		\bottomrule
	\end{tabular}
    \vspace{5pt}
	\label{tbl:weight}
\end{table}
\begin{table}[tbp]
	\centering
	\tiny
    \caption{
    The performance of LIMM on the AI-TOD dataset and the TinyPerson dataset. 
    The best performance is shown in \textbf{bold} and the second best is shown in \underline{underlined}. 
    ``-" indicates that public performance data is unavailable. 
    ``$\uparrow$" indicates that the larger the value, the better the performance.
    }
	\begin{tabular}[]{ccc}
		\toprule
        Dataset & AI-TOD & TinyPerson \\
        \midrule
		Method & $\rm AP_{50}$ $\uparrow$ & $\rm AP_{50}$ $\uparrow$ \\
		\midrule
        SAFF-SSD~\cite{huo2023self} (RS 23) & 49.9 & 52.5 \\
        MENet~\cite{zhang2024multistage} (TGRS 24) & \textbf{56.2} & 57.2 \\
		HANet~\cite{guo2024save} (TCSVT 24) & 53.7 & \underline{58.5} \\
		\midrule
		ConvNeXt-T~\cite{liu2022convnet} (CVPR 22) & 48.9 & 51.3 \\
		LIMM + ConvNeXt-T & \underline{54.2} & \textbf{58.8} \\
		\bottomrule
	\end{tabular}
	\label{tbl:tod}
\end{table}

\subsection{Visualization of Density Maps}
The visualization of the density maps generated by LIMM, the baseline ConvNeXt and the SOTA method MAN~\cite{lin2022boosting} is shown in Figure~\ref{fig:density_visual}. 
In the areas marked with red boxes, the baseline ConvNeXt and MAN shows significant foreground-background recognition errors, whereas our method does not exhibit such errors.
This advantage can be owed to the fact that the background regions with a density of 0 and the foreground regions with non-zero density are not in the same density level. 
The contrastive learning strategy guides the model to distinguish between them.
In the other images without red frames, ConvNeXt and MAN can correctly distinguish between the foreground and background, but its counting accuracy is far inferior to ours.
This indicates that ConvNeXt and MAN cannot accurately estimate the density of the crowd in the foreground area. 
Nevertheless, LIMM can make very precise density estimations because it only focuses on the content within the window, and contrastive learning forces the model to distinguish between different density levels.

\subsection{Computational Complexity Analysis}
Table~\ref{tbl:weight} shows the comparison of computational complexity and number of parameters between LIMM and several SOTA methods.
In general understanding, the window partition design introduces additional edge padding operations, which may increase the computational complexity.
However, the PyTorch framework has special optimizations for such zero-input operations, where neighboring zero input units will not activate their corresponding convolutional unit in the next layer.
Therefore, both the window partition and contrastive learning strategies do not introduce additional computational cost or parameters.
Compared to the baseline model, LIMM only introduces 0.5 GFLOPs of computational cost and 1M parameters due to the inclusion of the GSA layer.
In contrast, CLM~\cite{chen2024learning}, which is also a crowd counting method based on contrastive learning, introduces 1.5 GFLOPs of computational cost.
Compared to other SOTA methods, such as MAN and Gramformer, LIMM shows superior performance while requiring only half of their computational cost.
We can owe it to the fact that the small receptive field prior in our method helps avoid more unnecessary computations.

\subsection{Performance on Tiny Object Detection}
\label{sec:tiny_obj}
Though this paper focuses on designing crowd counting methods that align with the characteristic of small-sized individual's domination, crowd counting is not the only task characterized by small targets.
Tiny object detection (TOD) represents another typical example, which have been introduced in Section~\ref{sec:related_tod}.
Individuals in TOD are generally significantly smaller than those in crowd counting, and TOD datasets rarely contain large-sized objects.
Although the design of TOD methods differ slightly from crowd counting, we still apply LIMM to the TOD task to further validate its capability in detecting small-sized targets.
Here, our LIMM adopts ConvNeXt-T as the backbone and replaces the head with the same structure as Cascade R-CNN~\cite{cai2018cascade} to adapt to the object detection task.
The window size is set to 64 to adapt to smaller individual sizes in TOD.
We conduct experiments on the AI-TOD dataset~\cite{wang2021tiny} and the TinyPerson dataset~\cite{yu2020scale}. 
Table~\ref{tbl:tod} shows the results. 
On the AI-TOD dataset, LIMM demonstrates competitive performance, though slightly inferior to state-of-the-art methods.
We think the gap stems from task disparities.
For example, the GSA layer in LIMM fails to leverage its strengths in the TOD task.
In the TinyPerson dataset with slightly larger target sizes, LIMM achieves state-of-the-art performance.
Above experiments shows LIMM's broad potential for small-target visual tasks.

\section{Conclusion and Limitations}
This paper investigates the characteristics of crowd counting data and models.
A model design principle is proposed: emphasizing local modeling capability of the model.
Based on the principle, a simple yet effective method LIMM is designed. 
Experimental results validate the rationality of our proposed principle and the effectiveness of LIMM.
It is hoped that the exploration in this paper can provide new insights for crowd counting model design.
Nonetheless, there exists some limitations of our work.
For example, LIMM can be broadly applied to visual tasks dominated by small individuals, but this paper mainly focuses on its performance in crowd counting.
Exploring a universal paradigm for small-individual visual tasks is the direction for our future work.

\section*{Acknowledgements}
{
This research is partially supported by the National Natural Science Foundation of China (62176123, 62476130), and the Natural Science Foundation of Jiangsu Province (BK20242045).
}




\bibliography{mybibfile}

\appendix

\appendix
\section{Appendix Introduction}
In the appendix, we include additional content that is not covered in the main text due to space constraints.
In Section~\ref{sec:structure_of_heads}, the head structure of crowd counting and image classification is introduced.
The method for computing individual sizes in the histograms is introduced in Section~\ref{sec:computing_individual_sizes}.
In Section~\ref{sec:mqcl}, a detailed introduction of multi-queue contrastive learning is provided. 
An image masking experiment is presented in Section~\ref{sec:image_masking} to further validate the answer to Question 2. 
Subsequently, the optimal values of the window size and the number of density levels $N$ are explored on all the four datasets in Section~\ref{sec:effect_hyper}.
The t-SNE maps for validating the effectiveness of contrastive learning are presented and analyzed in Section~\ref{sec:tsne}.
Finally, Section~\ref{sec:dataset} provides an overview of all the datasets used in the paper.

\section{Structure of Heads}
\label{sec:structure_of_heads}
\begin{figure}[tbp]
	\centering
    \includegraphics[width=180pt]{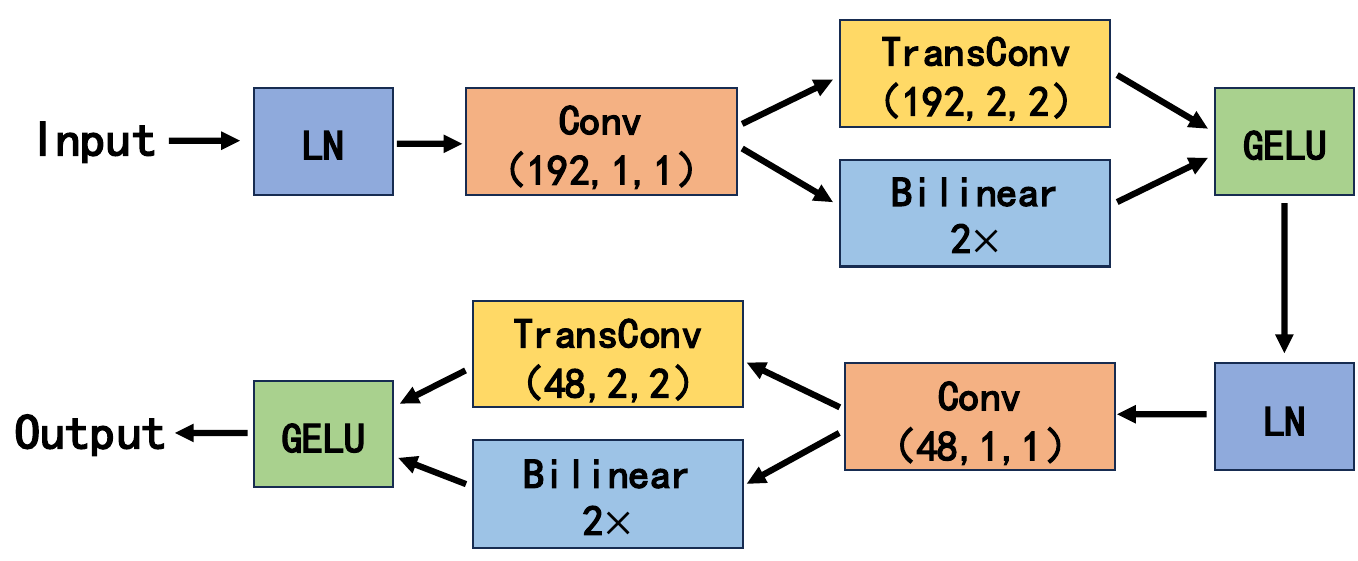}
	\caption[]{Structure of the crowd counting head. The convolutional and transposed convolutional layers' parameters are denoted as ``Conv/TransConv(number of kernels, kernel size, stride)". Bilinear 2$\times$ represents a bilinear interpolation layer which upsample the width and height to twice their original size. LN represents layer normalization and GELU represents GELU activation function. }
    \vspace{20pt}
	\label{fig:head_structure}
\end{figure}

The head structure of crowd counting is shown in Figure~\ref{fig:head_structure}. 
Its core are two upsampling layers, which enables the generation of higher-resolution density maps.
The upsampling layers are dual-branch structures, incorporating both a transposed convolution and a bilinear interpolation.
The feature maps are added after passing through them.
The introduction of this dual-branch structure is intended to avoid the checkerboard artifacts caused by using a single transposed convolution.
The structure of the image classification head remains consistent with the original ConvNeXt~\cite{liu2022convnet}, consisting of a global average pooling layer followed by a fully connected layer.
The structure of the object detection head is consistent with that of Cascade R-CNN~\cite{cai2018cascade}.

\section{Method for Computing Individual Sizes}
\label{sec:computing_individual_sizes}
Here we specifically elaborate how the sizes of individuals are computed in Figure 2 of the main text. 
Some of the datasets in Figure 2 in the main text provides bounding box annotations, such as JHU-Crowd++~\cite{sindagi2020jhu}, ImageNet-1K~\cite{deng2009imagenet} and COCO~\cite{lin2014microsoft}.
We can calculate the size $s$ of each individual using the formula: $s = (h \times w)^{0.5}$, where $h$ and $w$ is the height and width of the bounding box, respectively. 
For other crowd counting datasets that only provide point annotations, we follow the head size estimation method outlined in MCNN~\cite{zhang2016single}: taking the average distance to the three nearest neighbors.

\section{Multi-queue Contrastive Learning}
\label{sec:mqcl}
Multi-queue contrastive learning~\cite{pan2024boosting}, which is a representation learning method for unbalanced multi-class scenarios, is first proposed to handle the adverse weather crowd counting task.
First, the representation $R \in \mathbb{R}^{H \times W \times C_{1}}$ (the output of the encoder) is processed by projection heads $P_{Q}$ and $P_{K}$, generating 1-D vector $Q_{i}$ and $K_{i}$, respectively. 
These two projection heads share the same structure but have different parameters.
The projection head first pools the representations to vectors of $\mathbb{R}_{C_{1}}$ and then project them to vectors $V_{Q}$ or $V_{K}$ of $\mathbb{R}_{C_{2}}$ by introducing a multi-layer perceptron.
$C_{2}$ is the dimension of the vectors.
The multi-queue structure is introduced to store the vectors $V_{K}$.
Each of its sub-queues corresponds to a category (e.g., in this paper, a sub-queue corresponds to a density level).
The multi-queue structure can be considered as a tensor of $\mathbb{R}^{B \times L \times C_2}$, where $B$ is the number of classes (e.g., in this paper, $B$ is the number of density levels $N$ and $L$ is the length of each sub-queue. 
Immediately when the computation of vector $V_{Ki}$ is completed, it will be pushed into the corresponding sub queue according to its class. 
when a batch of data is processed, the loss function can be calculated as:
\begin{equation}
    L_{C} = \frac{1}{|N_{Q}|} L_{i}, 
    \label{eq:lc}
\end{equation}
where $N_{Q}$ is the size of the batch and $L_{i}$ is the loss corresponding to the vector $V_{Qi}$: 
\begin{equation}
    L_{i} = \frac{-1}{|P_{i}|} \sum_{p \in P_{i}} log{\frac{exp(s_{i,p} / \tau)}{\sum_{j=1}^{J} exp(s_{i,j} / \tau)}}, 
\end{equation}
where $P_{i}$ is the set of samples in the queue corresponding to the class of $V_{Qi}$, $s_{i,p}$ is the cosine similarity between $V_{Qi}$ and $V_{Kp}$. 
Through above learning process, the representations encoded by the encoder will be closer for the same category, while those for different categories will be relatively farther apart.
\begin{figure}[tbp]
	\centering
    \includegraphics[width=160pt]{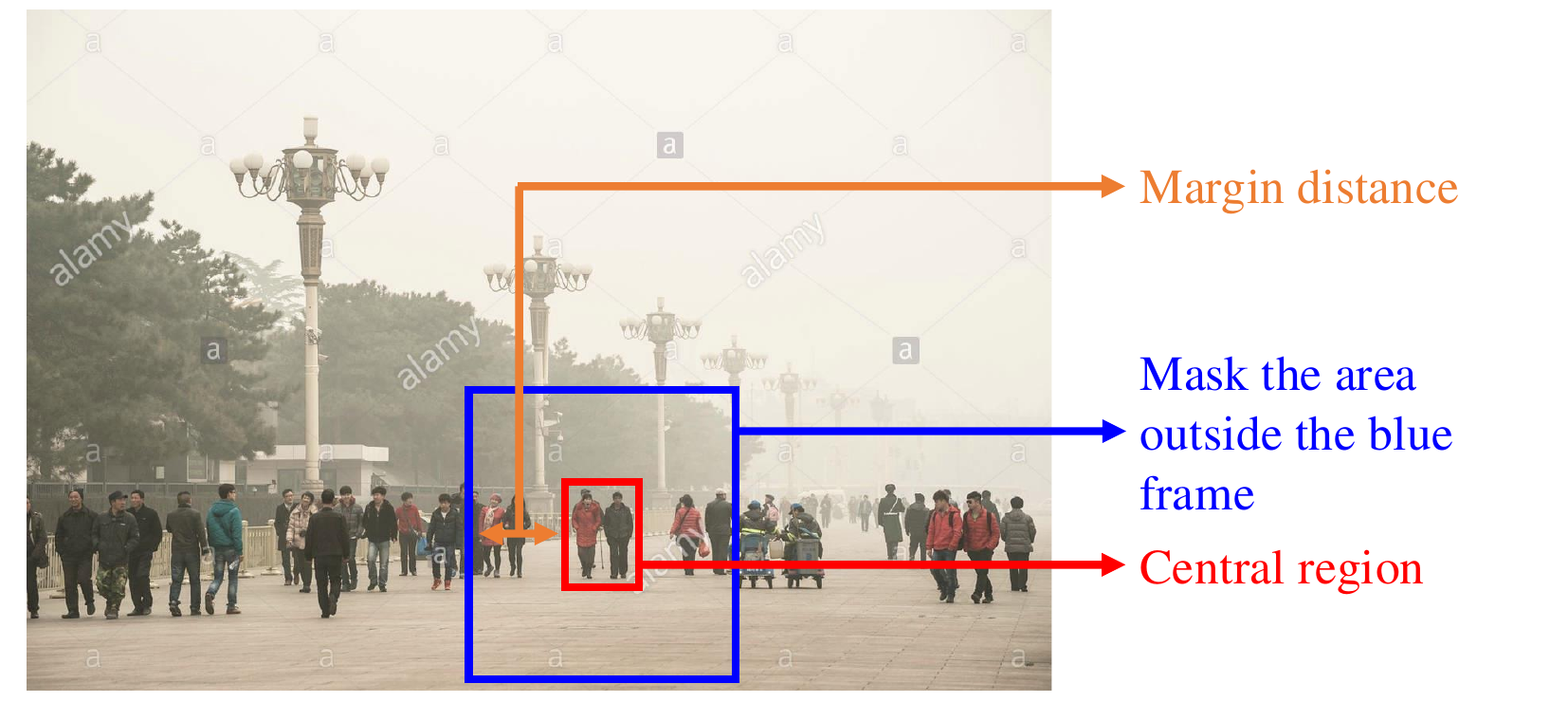}
	\caption[]{Illustration for image masking method. }
    \vspace{20pt}
	\label{fig:mask_strategy}
\end{figure}
\begin{figure}[tbp]
	\centering
    \includegraphics[width=235pt]{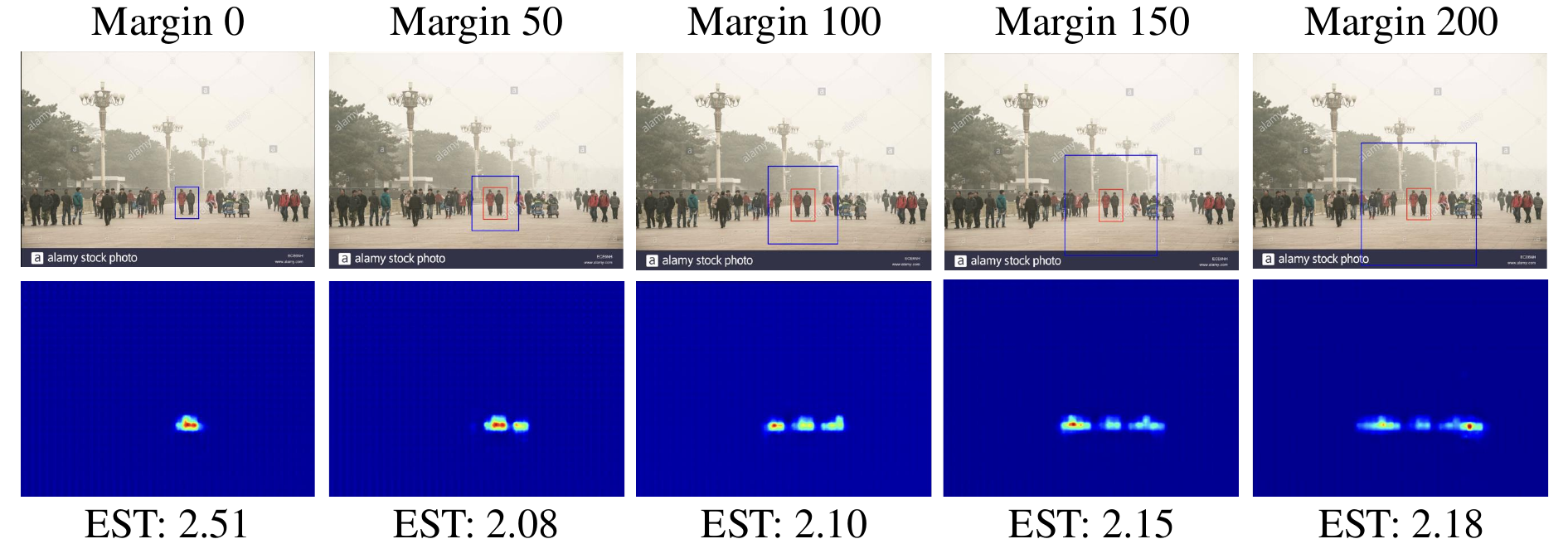}
	\caption[]{Visualization of image masking experiment. The image is selected from the JHU-Crowd++ dataset and the model is ConvNeXt-T trained on the JHU-Crowd++ dataset. }
    \vspace{20pt}
	\label{fig:mask_visual}
\end{figure}
\begin{figure*}[tbp]
	\centering
    \subcaptionbox{ShanghaiTech A}{
        \includegraphics[height=90pt]{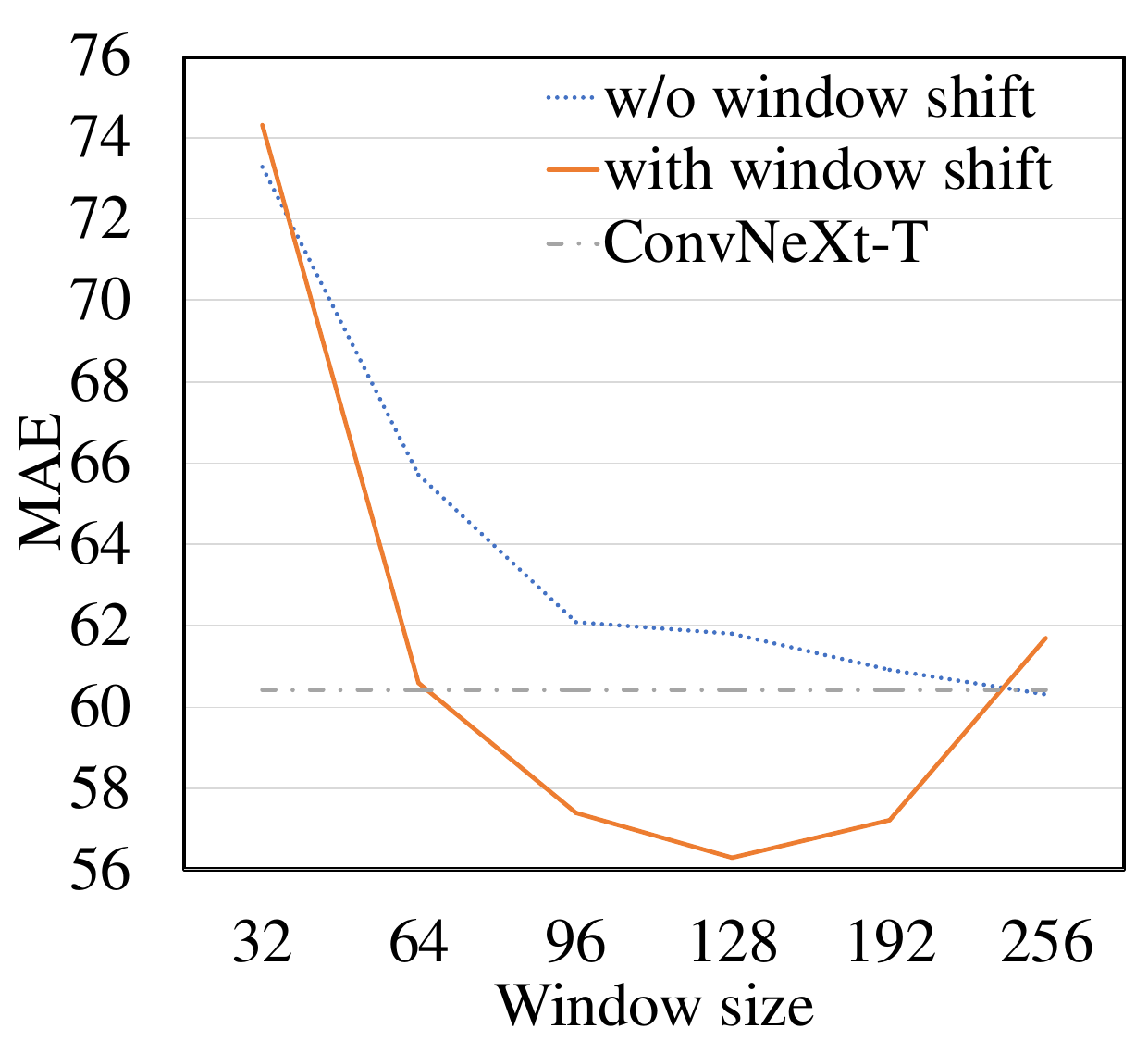}
    }
    \subcaptionbox{ShanghaiTech B}{
        \includegraphics[height=90pt]{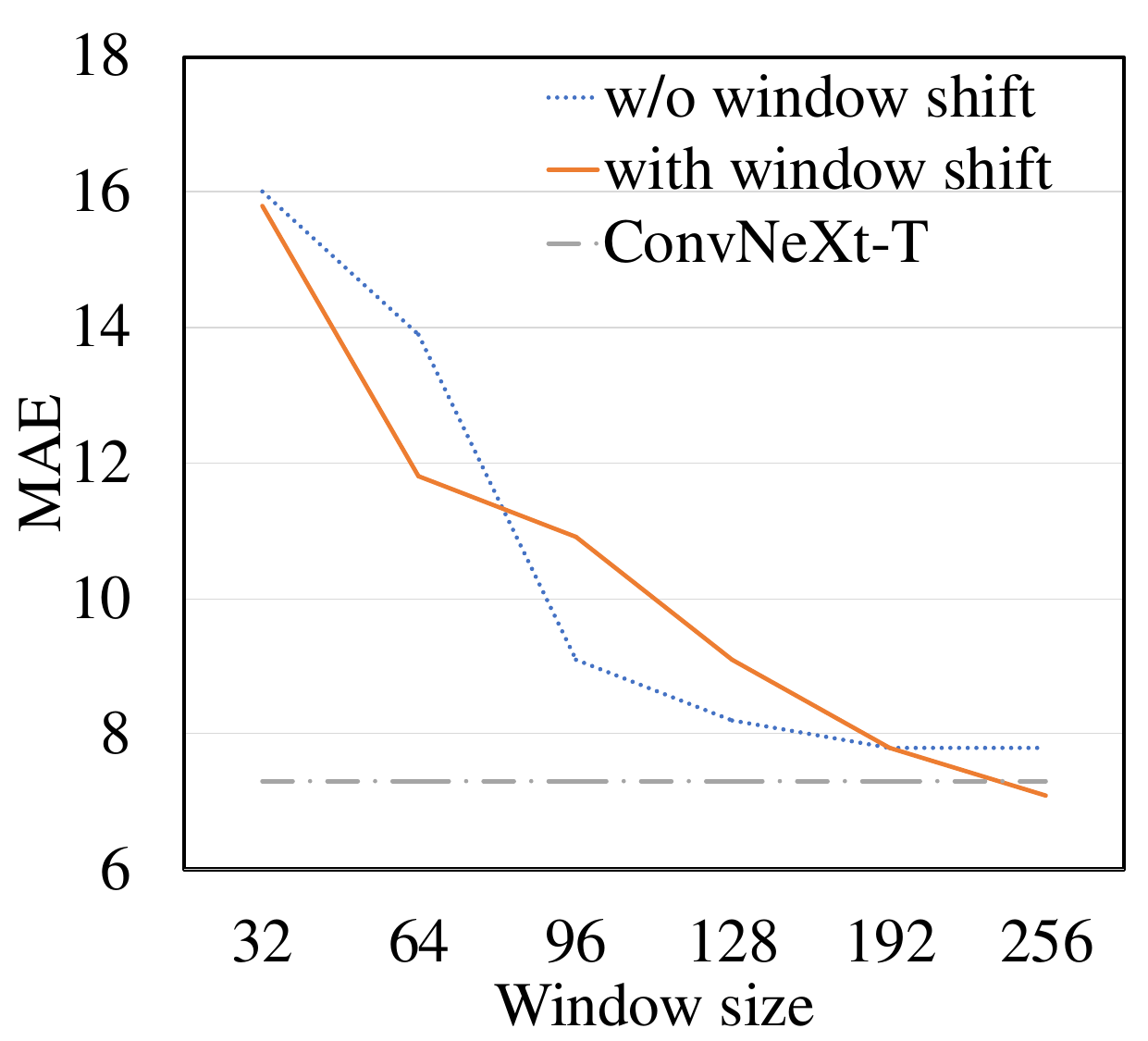}
    }
    \subcaptionbox{UCF-QNRF}{
        \includegraphics[height=90pt]{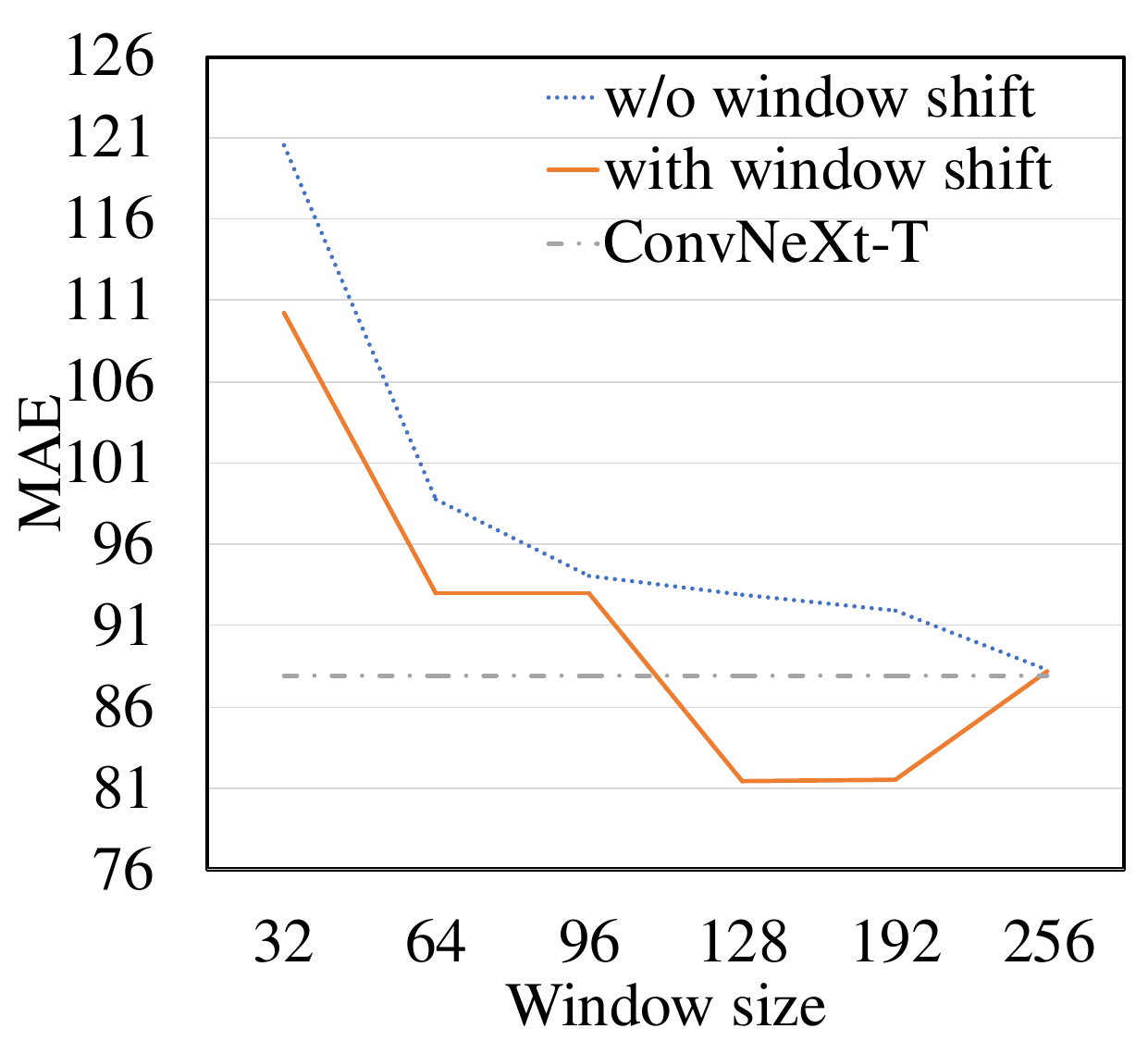}
    }
    \subcaptionbox{JHU-Crowd++}{
        \includegraphics[height=90pt]{window_size.pdf}
    }
    \vspace{10pt}
	\caption[]{
 Effect of window size on different datasets. 
 }
	\label{fig:window_size_all}
\end{figure*}
\begin{figure*}[tbp]
	\centering
    \subcaptionbox{ShanghaiTech A}{
        \includegraphics[height=90pt]{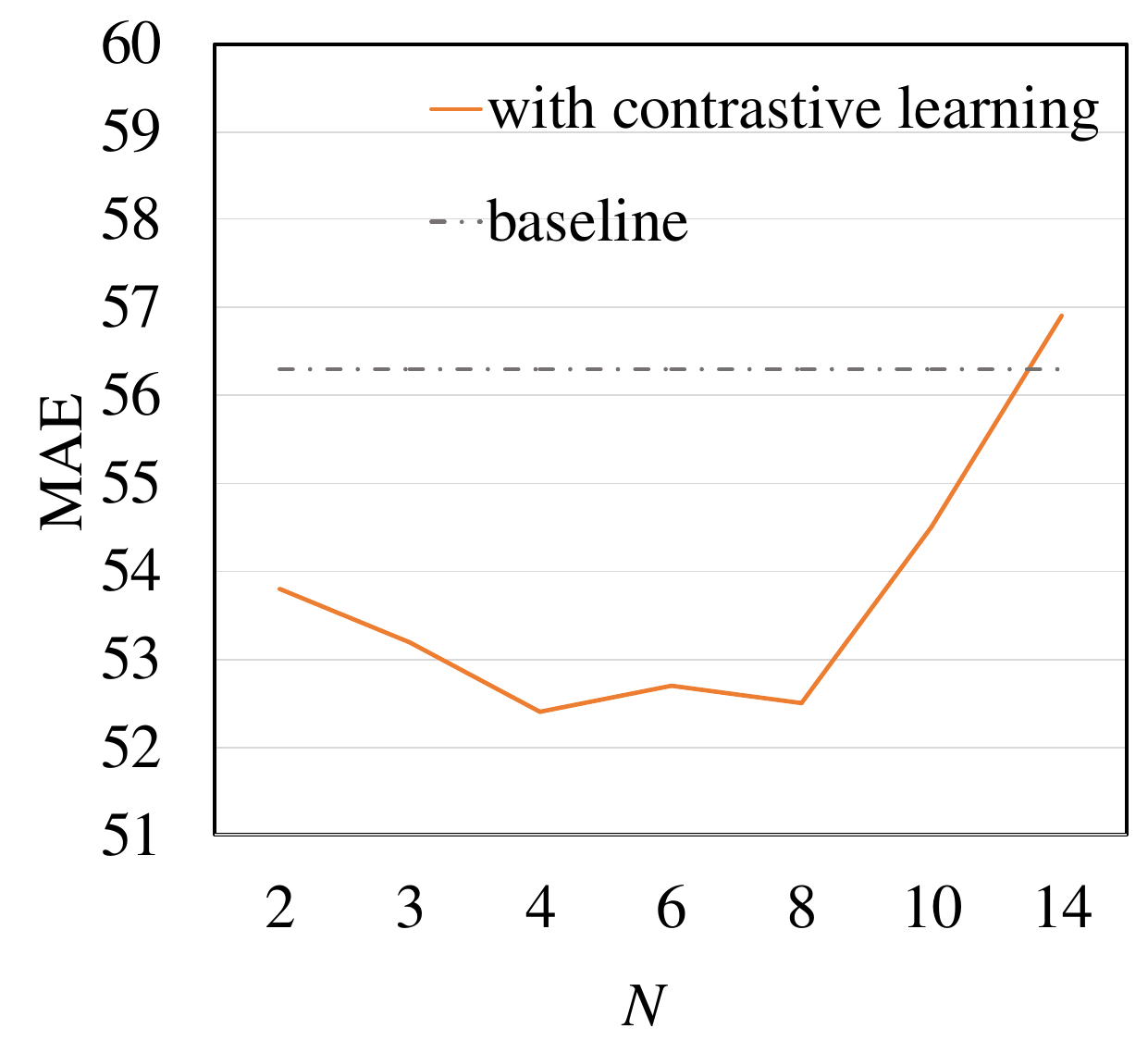}
    }
    \subcaptionbox{ShanghaiTech B}{
        \includegraphics[height=90pt]{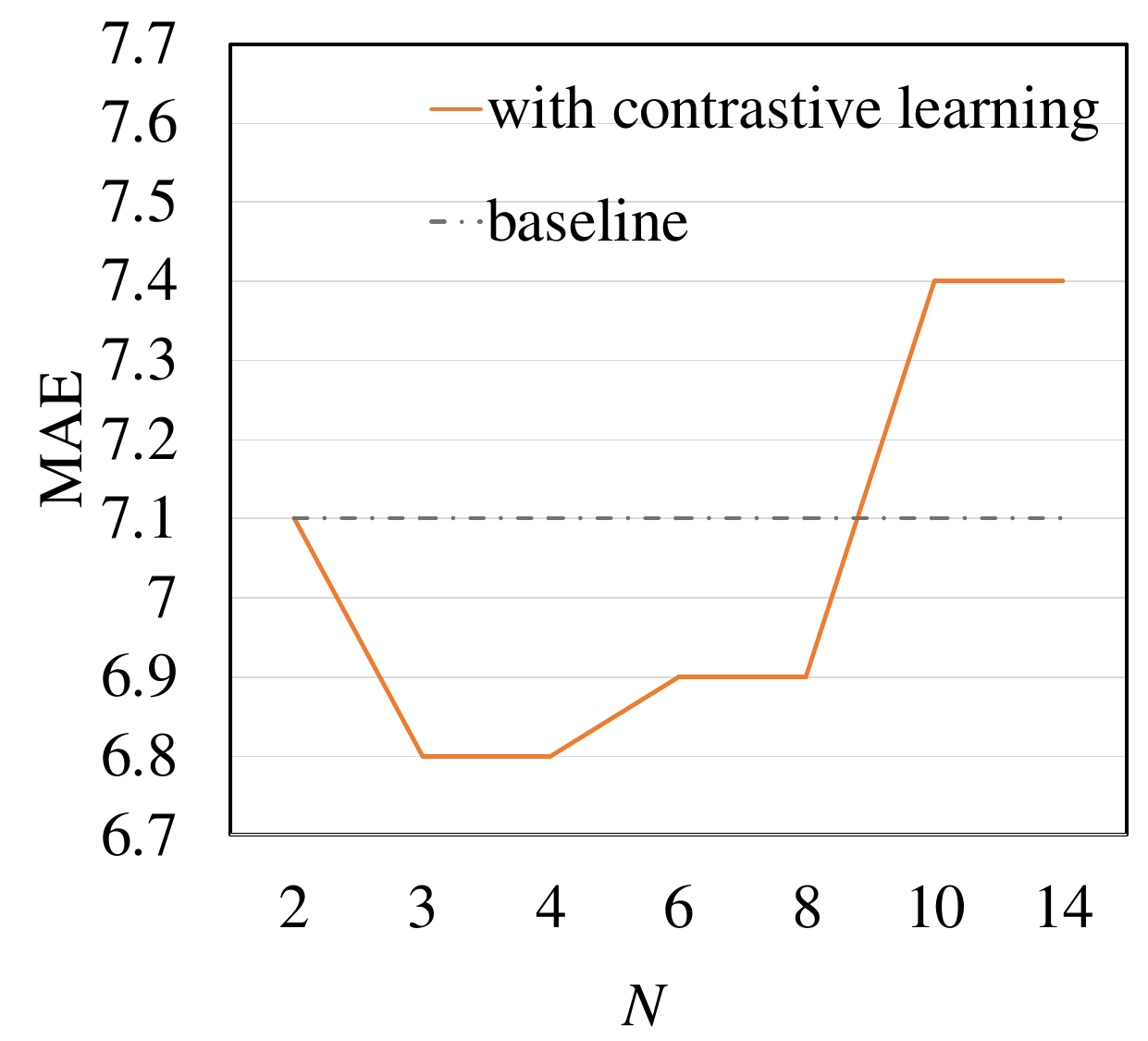}
    }
    \subcaptionbox{UCF-QNRF}{
        \includegraphics[height=90pt]{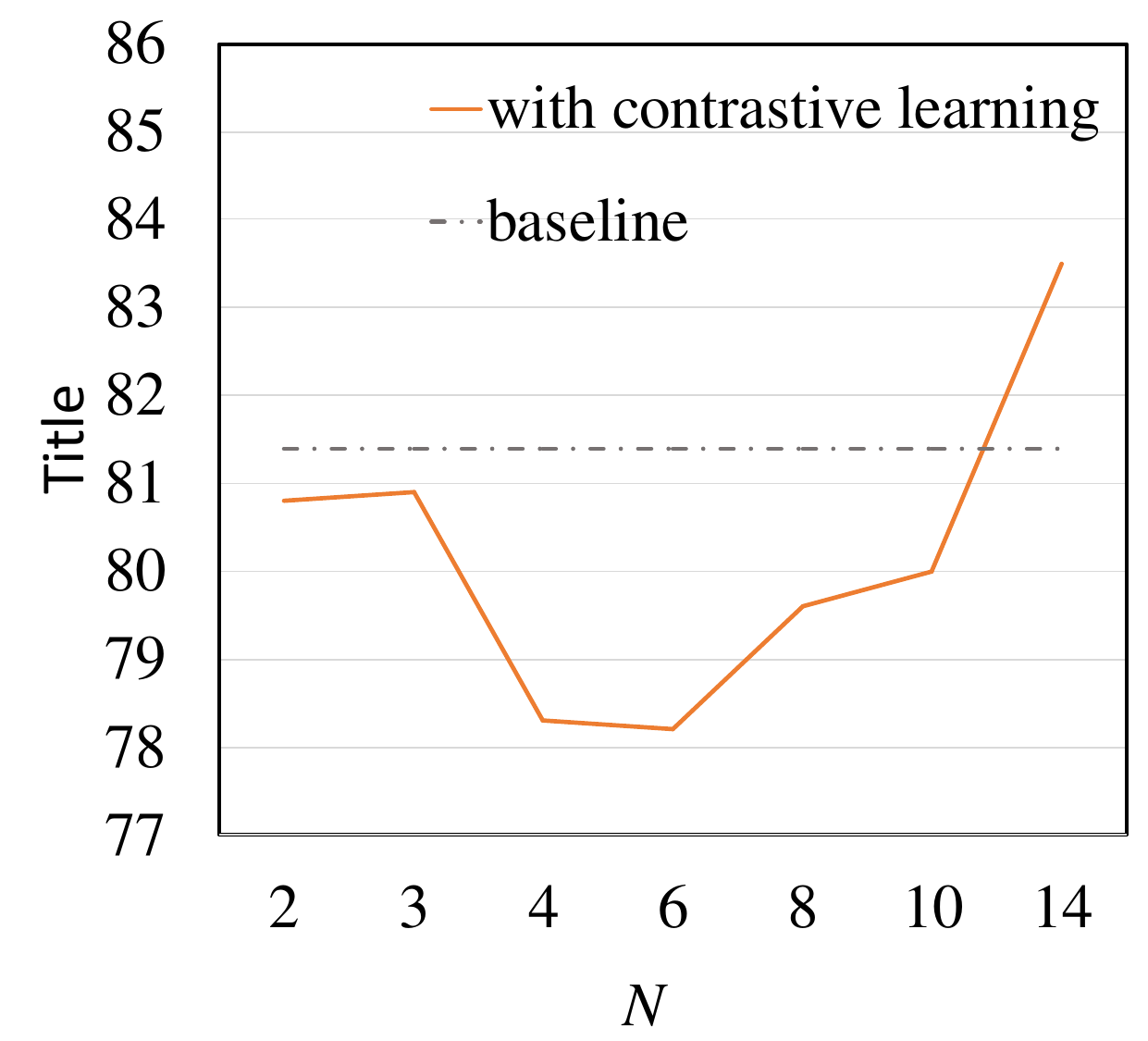}
    }
    \subcaptionbox{JHU-Crowd++}{
        \includegraphics[height=90pt]{density_level.pdf}
    }
    \vspace{10pt}
	\caption[]{
 Effect of the number of density levels $N$ on different datasets.
 The baseline is ConvNeXt-T with the window partition design. 
 }
 \vspace{10pt}
	\label{fig:density_level_all}
\end{figure*}
\begin{figure}[tbp]
	\centering
    \includegraphics[width=200pt]{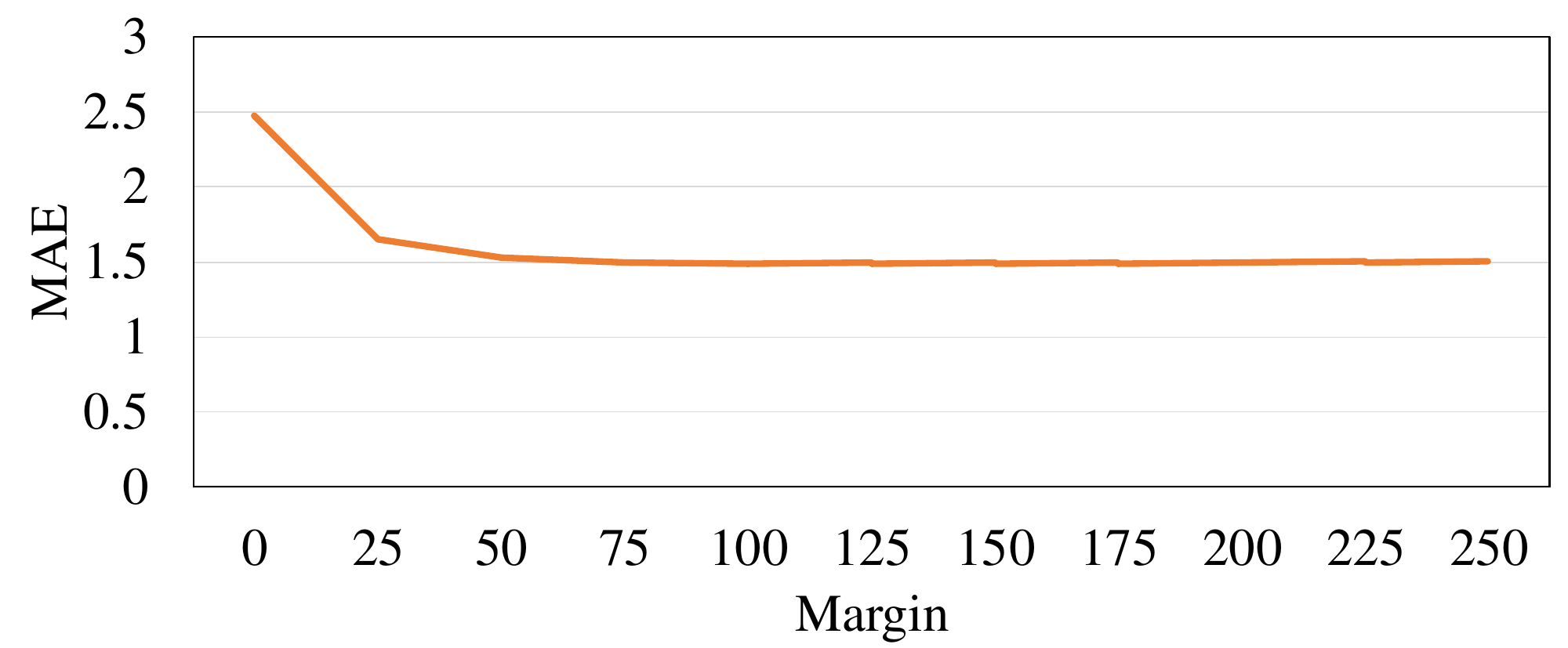}
	\caption[]{Line chart of quantitative results for image masking experiment. The experiment was conducted on the test set of the JHU-Crowd++ dataset and the model is ConvNeXt-T trained on the JHU-Crowd++ dataset. }
	\label{fig:mask_quantify}
    \vspace{15pt}
\end{figure}

\section{Image Masking Experiment}
\label{sec:image_masking}
In addition to ERF, image masking is another effective method to investigate the attention scope of crowd counting models. 
The illustration of the image masking experiment is shown in Figure~\ref{fig:mask_strategy}.
Our specific strategy is to select a region in the image as the central region and set a certain number of pixels, referred to as the margin distance.
The image within this distance from the central region are retained, while all pixels outside this distance are set to 0.
Then, we gradually increase the margin distance and observe the trend in the model's output value for the central region.
If the output value changes significantly, it indicates that the surrounding area of the image has a substantial impact on the model output; conversely, if the output value remains relatively stable, it suggests that the model does not consider regions beyond the local area.
We still use ConvNeXt-T trained on the JHU-Crowd++ dataset for our experiments. 
An example is visualized in Figure~\ref{fig:mask_visual}.
When the margin distance is small, such as within 50 pixels, the model's estimation of the central region fluctuates due to the boundary effect.
However, as the edge distance continues to increase, the model's estimation stabilizes.
This indicates that the model focuses more on the content near the central region itself rather than on distant information.
Additionally, quantitative experiments are also conducted on the whole JHU-Crowd++ test set. 
On each image from the JHU-Crowd++ test set, 5 central regions with the size of 100 $\times$ 100 are randomly sampled.
As the margin distance gradually varies from 0 to 250, the average difference between the model's output and the ground truth number of the central region are statistically analyzed. 
The line chart is shown in Figure~\ref{fig:mask_quantify}. 
The results are consistent with the visualization, indicating that image content more than 50 pixels away from the central region has no impact on the prediction results.
The image masking experiment confirms that the attention scope of crowd counting models is usually quite small.

\section{Effect of Hyperparameters}
\label{sec:effect_hyper}
Although we have preliminarily explored the values of the hyperparameters in the preceding discussion, to further explore the characteristics of the window size and the number density levels $N$, we investigate their optimal values on all the four datasets, as shown in Figure~\ref{fig:window_size_all} and \ref{fig:density_level_all}. 
Across most datasets, their optimal configurations consistently align with the values of 128 and 6, which has been mentioned in the main text.
The sole exception occurs in the ShanghaiTech Part B dataset, where the optimal window size increases. 
This deviation stems from the dataset's higher proportion of large-sized individuals.
Additionally, the characteristic of diminished size variation enables the model to achieve satisfactory performance in window-wise contrastive learning with a relatively small number of density levels.
In conclusion, unless dealing with highly atypical data, the optimal values for these two hyperparameters remain relatively stable.

\section{Visualization of Contrastive Learning}
\label{sec:tsne}
\begin{figure}
	\centering
    \subcaptionbox{LIMM w/o DW}{
        \includegraphics[height=80pt]{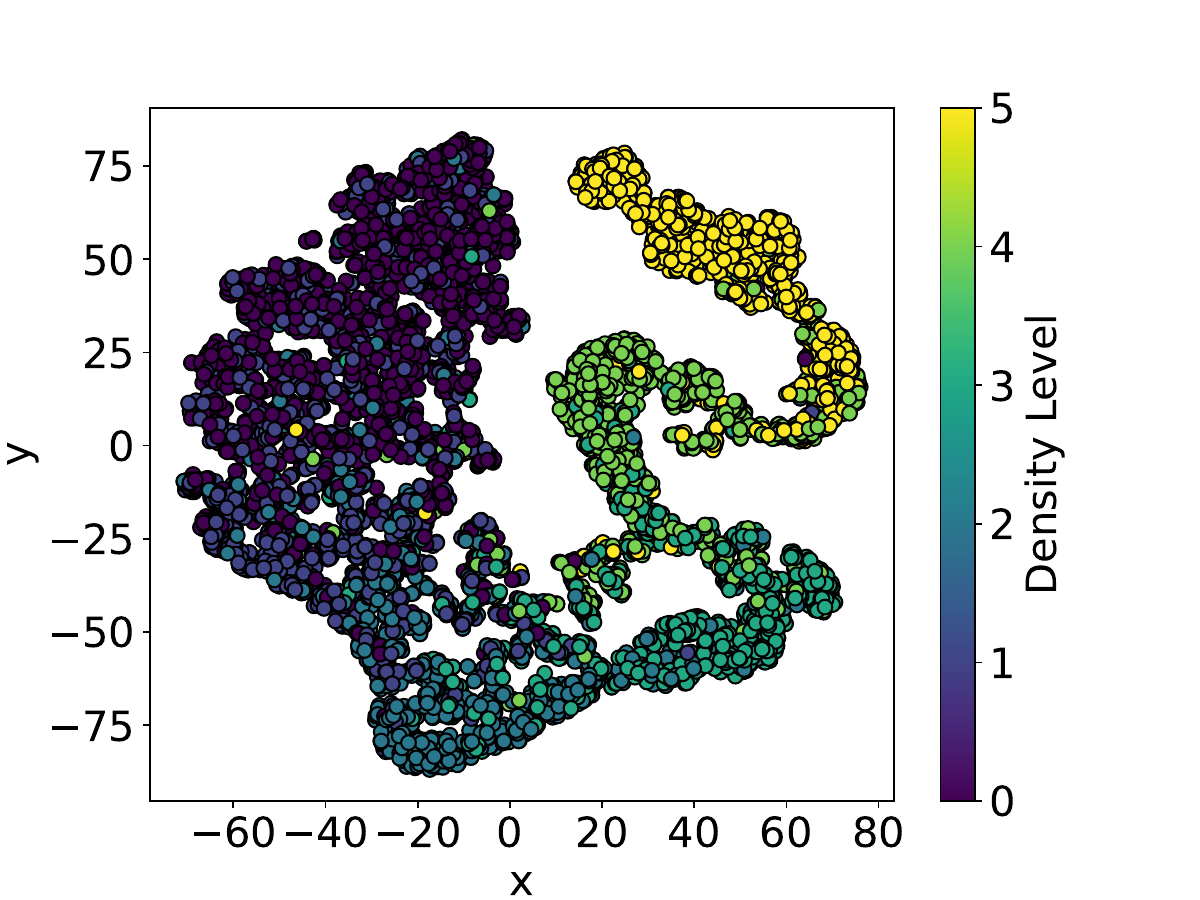}
    }
    \subcaptionbox{LIMM}{
        \includegraphics[height=80pt]{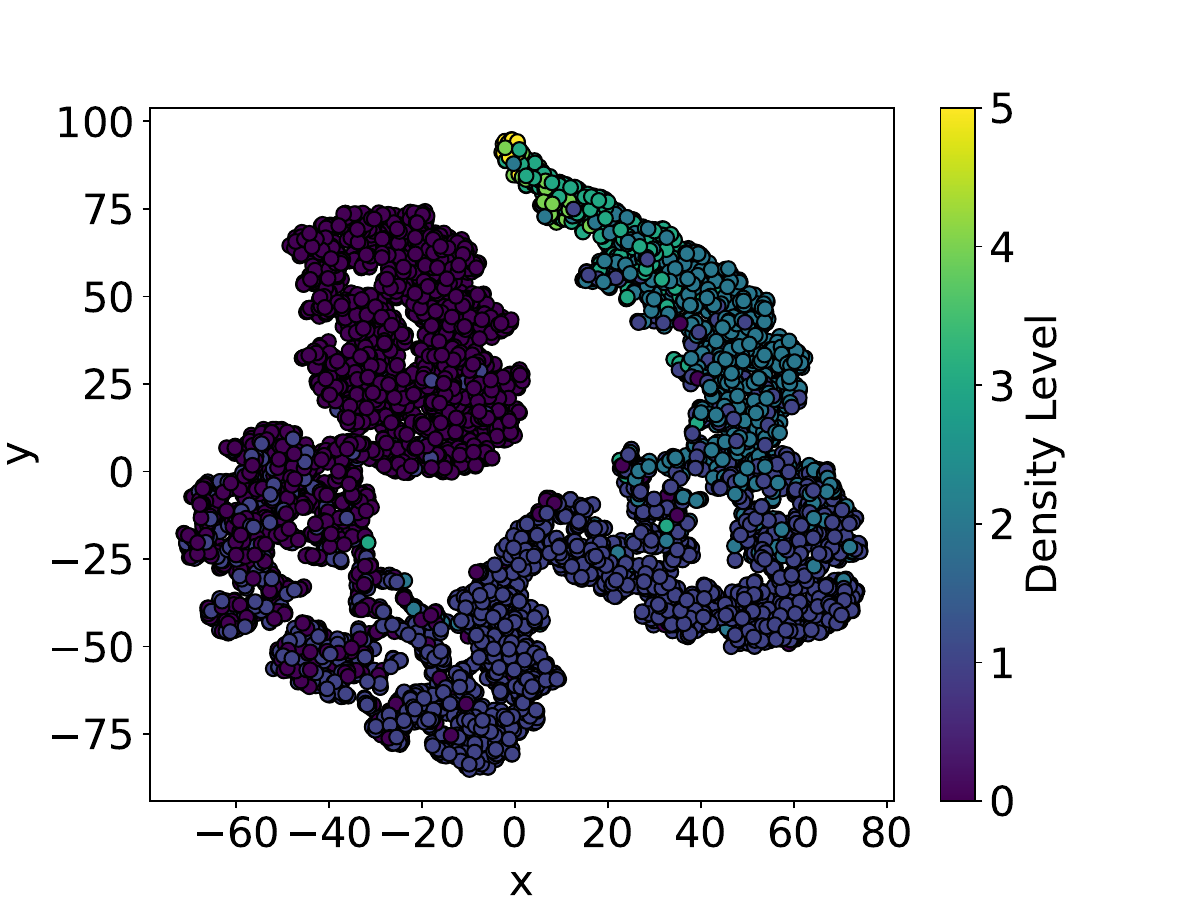}
    }
    \vspace{10pt}
	\caption[]{
 The t-SNE visualization of the vectors $V_{Q}$ of LIMM (b) and LIMM without class distance weighting design (a). 
 }
    \vspace{20pt}
	\label{fig:tsne}
\end{figure}

To validate the effctiveness of the window-wise contrastive learning, we sample five windows from each image in the JHU-Crowd++~\cite{sindagi2020jhu} test set and calculate their corresponding vectors $V_{Q}$ to plot the t-SNE~\cite{van2008visualizing} scatter map.
Figure~\ref{fig:tsne} (a) corresponds to LIMM without the distance weighting design, while Figure~\ref{fig:tsne} (b) represents the full version of LIMM.
The color of the scatter points corresponds to different density levels.
The results show that both methods can effectively distinguish between different density levels. 
However, the full version of LIMM, by applying varying degrees of repulsion to negative samples at different density distances, achieves higher cohesion within the same category and greater separation between different categories.

\section{Descriptions of Datasets}
\label{sec:dataset}
\textbf{ShanghaiTech.}
The ShanghaiTech dataset~\cite{zhang2016single} contains two parts (A and B). 
Part A is sourced from various locations around the world and its density is high, which contains 482 images with 244,167 annotated points.
300 images are divided for training and the remaining 182 images are for testing.
The images in Part B are sourced from Shanghai, where the crowd density is relatively low.
On average, the heads of humans in Part B occupy a larger proportion of the image.
400 images of Part B are for training and 316 for testing. 
\noindent
\textbf{UCF-QNRF.}
The UCF-QNRF dataset~\cite{idrees2018composition} includes 1,535 high resolution images collected from the Web, with 1.25 million annotated points. 
There are 1,201 images in the training set and 334 images in the testing set.
UCF-QNRF is extremely high-density and the range of people count is wide. 
\noindent
\textbf{JHU-Crowd++.}
There are 4372 images and 1.51 million labels contained in the JHU-Crowd++ dataset~\cite{sindagi2020jhu}. 
The images are collected from several sources on the Internet using different keywords and specifically chosen for adverse weather conditions.
Out of these, 2272 images were used for training, 500 images for validation, and the remaining 1600 images for testing.
The advantage of JHU-Crowd++ is its inclusion of high-density scenes and diverse environmental conditions.
%


\end{document}